\pdfoutput=1

\documentclass[11pt]{article}

\usepackage[final]{acl}

\usepackage{times}
\usepackage{latexsym}

\usepackage[T1]{fontenc}

\usepackage[utf8]{inputenc}

\usepackage{microtype}

\usepackage{inconsolata}
\usepackage{soul}
\usepackage{color}
\usepackage{xcolor}
\usepackage{float} 
\usepackage{subfig}
\usepackage{overpic}     
\usepackage{amsmath}
\usepackage{stfloats}
\usepackage{multirow}
\usepackage{enumitem}
\usepackage{booktabs}
\usepackage{bbding}
\usepackage{diagbox}
\usepackage{newfloat}
\usepackage{listings}
\usepackage{courier}  
\usepackage{graphicx}
\usepackage{natbib}  
\usepackage{caption}
\usepackage{makecell}
\usepackage{algorithm}
\usepackage{algorithmic}
\usepackage{wrapfig}
\usepackage{colortbl}
\definecolor{lightgreen}{RGB}{202,252,209}
\definecolor{generative}{RGB}{232, 241, 238} 
\definecolor{retrieval}{RGB}{250, 229, 215} 
\definecolor{pretraining}{RGB}{255, 242, 206} 
\title{Improving Large Language Models in Event Relation Logical Prediction}

\author{
 Meiqi~Chen$^{1}$\footnotemark[1], Yubo~Ma$^{2}$, Kaitao~Song$^{3}$$\footnotemark[2]$, Yixin~Cao$^{4}$, Yan~Zhang$^{1}$$\footnotemark[2]$, Dongsheng~Li$^{3}$ \\ 
 $^1$ Peking University
 $^2$ Nanyang Technological University
 $^3$ Microsoft Research Asia \\
 $^4$ School of Computer Science, Fudan University
 \\
\texttt{meiqichen@stu.pku.edu.cn}, \texttt{yubo001@e.ntu.edu.sg}\\
\texttt{\{kaitaosong, dongsli\}@microsoft.com}, \\
\texttt{caoyixin2011@gmail.com}, \texttt{zhyzhy001@pku.edu.cn}
}
\begin{document}
\maketitle
\begin{abstract}
Event relations are crucial for narrative understanding and reasoning. Governed by nuanced logic, event relation extraction~(ERE) is a challenging task that demands thorough semantic understanding and rigorous logical reasoning.
In this paper, we conduct an in-depth investigation to systematically explore the capability of LLMs in
understanding and applying event relation logic. More in detail, we first investigate the deficiencies
of LLMs in logical reasoning across different tasks. Our study reveals that LLMs are not logically consistent reasoners, which results in their suboptimal performance on tasks that need rigorous reasoning. To address this, we explore three different approaches to endow LLMs with event relation logic, and thus enable them to generate more coherent answers across various scenarios. Based on our approach, we also contribute a synthesized dataset (\texttt{LLM-ERL}) involving high-order reasoning for evaluation and fine-tuning. Extensive quantitative and qualitative analyses on different tasks also validate the effectiveness of our approaches and provide
insights for solving practical tasks with LLMs in future work. Codes are available at \url{https://github.com/chenmeiqii/Teach-LLM-LR}.
\end{abstract}

\newcommand{\tabincell}[2]{\begin{tabular}{@{}#1@{}}#2\end{tabular}}
\renewcommand*{\thefootnote}{\fnsymbol{footnote}}
\footnotetext[1]{This work was done during her internship at Microsoft Research Asia.}
\footnotetext[2]{Corresponding author.}
\section{Introduction}
\label{sec:intro}
\begin{figure}[htbp!]
\centering  
\includegraphics[width=0.48\textwidth]{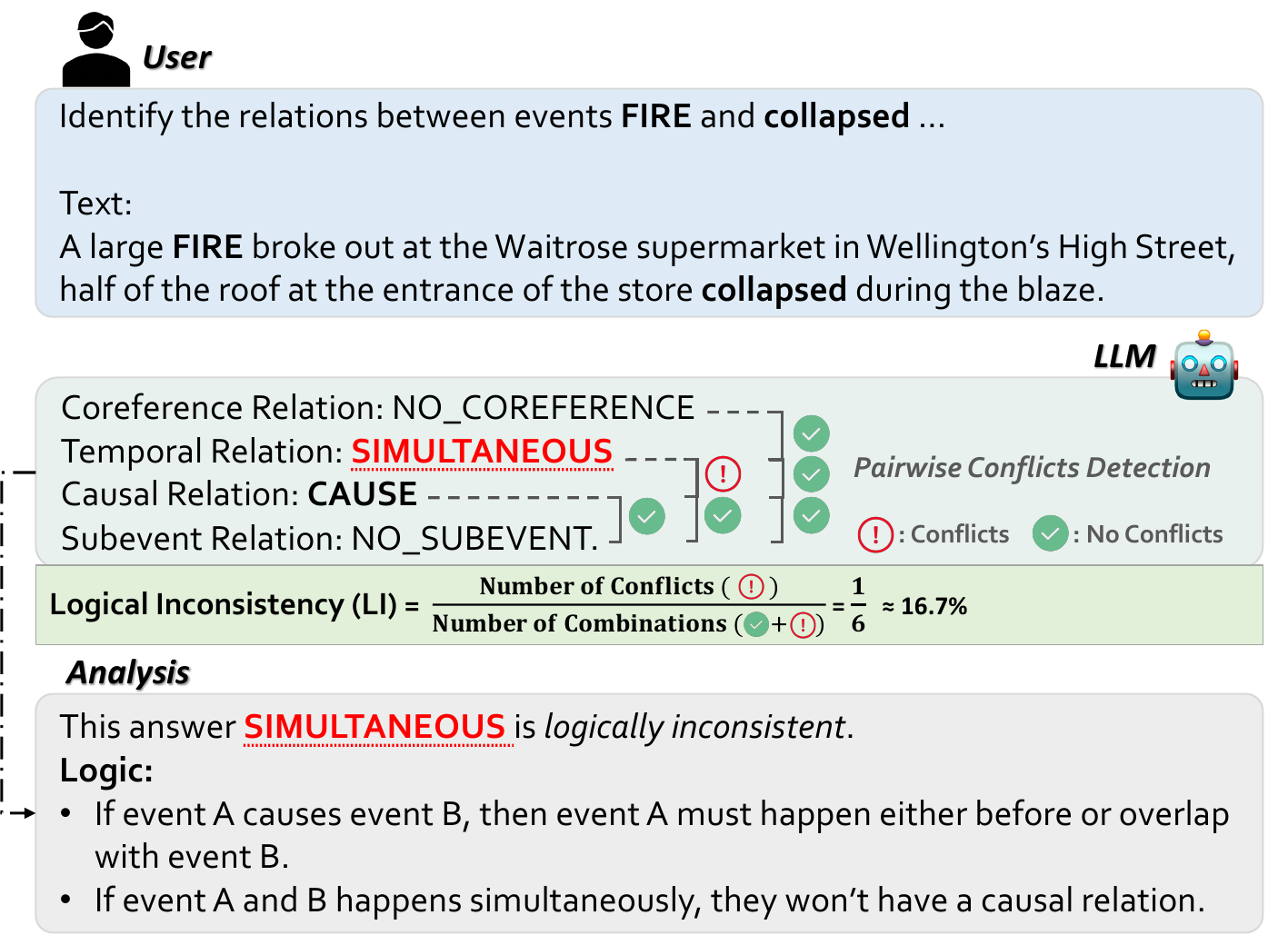} 
\caption{An example of LLM in generating logically inconsistent answers. We let an LLM (e.g., ChatGPT) predict the relations between events \emph{``FIRE''}  and \emph{``collapsed''} from the given passage. We can find that LLM predicts an incorrect answer (i.e., SIMULTANEOUS) because it ignores some prior logic in this scenario.}
\label{fig:example}
\end{figure}
Understanding the relationships between events is fundamental to effective communication and reasoning, a challenge central to the field of Event Relation Extraction (ERE). ERE tasks, which involve identifying coreference, temporal, causal, and subevent relationships, demand not only semantic comprehension but also rigorous logical reasoning. Despite recent advances in Large Language Models (LLMs) such as ChatGPT~\citep{ouyang2022training} and Llama2~\citep{touvron2023llama}, these models struggle to fully grasp the complexities of event relation logic, often failing to apply it accurately in ERE tasks.

As showcased in Figure~\ref{fig:example}, ChatGPT incorrectly predicts the temporal and causal relations between events \emph{``FIRE''}  and \emph{``collapsed''} as \emph{``simultaneous''} and \emph{``cause''}, respectively. According to the prior logical constraints, we could readily claim the predictions are not fully correct even before reading the context.
Some works~\citep{xu2023large, Liangming2023LogicLM, Qing2023FaithFulCoT} attribute this gap in logic adherence to LLMs' inherent deficiencies (e.g., hallucination, unfaithfulness). However, how to disentangle and improve the capability of LLMs in these tasks is still an open problem.

To deeply understand the deficiencies of LLMs in logical reasoning and explore the corresponding solutions, in this paper, we conduct an in-depth investigation of LLMs in solving reasoning tasks from multiple dimensions. Our experimental results show that:
\textbf{1)} Even the cutting-edge LLMs still generate large amounts of inconsistent answers, e.g., over 60\% of the answers from ChatGPT on the MAVEN-ERE~\cite{wang2022maven} dataset are logically inconsistent as shown in Figure~\ref{fig:lce}; \textbf{2)} Providing relevant logic to LLMs improves performance, but injecting irrelevant logic introduces fluctuations in results. Therefore, how to obtain the relevant logic and inject its information into LLMs is a non-trivial problem, deserving further exploration.

Based on these findings, we put forward a series of solutions to endow LLMs with event relation logic and generate more coherent answers. Here, we propose three different kinds of approaches according to the ways of logic acquisition:
\textbf{1)} \emph{Generative-based approach}, which encourages LLMs to generate rationale themselves, inspired by CoT prompting~\citep{Jason2022CoT}. In this paradigm, we find that incorporating logical constraints into LLM instruction will bring substantial improvements, but the uncertainty of the generated rationales may also bring some biases, leading to an incorrect subsequent answer; 
\textbf{2)}  \emph{Retrieval-based approach}, which collects constraints from realistic data, then retrieves relevant contents and adds them to the LLM instruction. This kind of approach ensures the correctness of logic and significantly improves performance, but requires some hand-crafted engineering; 
\textbf{3)} \emph{Finetuning-based approach}, which first constructs a high-order event relation logical prediction dataset (\texttt{LLM-ERL}), then uses it to fine-tune specialized LLMs. The finetuning dataset consists of multi-hop event relation logical prediction instances. This strategy encodes logic in model parameters inherently, making them more suitable for white-box LLMs.
Therefore, how to choose the most suitable strategy can be a trade-off based on the practical scenario.

Furthermore, based on the above framework, we also conduct extensive quantitative and qualitative analyses to validate the effectiveness of the proposed approaches and provide insights for future work: 
\textbf{1)} Directly using CoT to infer ERE tasks is limited by the inherent issues of LLMs, but incorporating logical constraints in the reasoning process can be beneficial;
\textbf{2)} Retrieval-based approaches can significantly reduce inconsistencies in LLM responses. Stronger models like GPT-4 can effectively perform retrievals by themselves, whereas weaker models require assistance in filtering relevant information. Besides, directly conveying constraints to LLMs is more effective than adding post-processing operations based on the results;
\textbf{3)} When fine-tuned on \texttt{LLM-ERL}, LLMs such as Llama2-13B~\citep{touvron2023llama} can achieve better performance, which validates the effectiveness of our proposed approaches.

Overall, the contributions of our paper can be summarized as follows:
\begin{itemize}[leftmargin=*]
\item We provide an in-depth investigation of the logical inconsistency issue of current LLMs, highlighting their challenges in understanding event relation logic.
\item We propose several solutions to endow LLMs with event relation logic and generate more coherent answers. Based on our approach, we construct a synthesized dataset (\texttt{LLM-ERL}) involving high-order reasoning to enhance LLMs. 
\item Experimental results on different tasks with quantitative and qualitative analyses further verify the effectiveness of our approach in endowing LLMs with event relation logic.
\end{itemize}

\section{Event Relation Logic}

\subsection{Event Relations}
In this subsection, we introduce four common types of event
relations that are crucial for narrative comprehension and reasoning. 
\emph{Coreference relations}: identify whether two event mentions refer to the same occurrence. 
\emph{Temporal relations}: establish the chronological order of events.
\emph{Causal relations}: identify causality between events.
\emph{Subevent relations}: identify whether one event is a subcomponent of another.
More descriptions of these event relations can be found in Appendix~\ref{app:temp}.

Based on these four relations,  event relation extraction (ERE) can be formulated as a multi-label classification problem, assigning one label for each relation type. Compared with other common tasks, ERE tasks should take more considerations about the logical constraints between event relations (e.g., as shown in Figure~\ref{fig:example}), and guarantee the predictions should conform to these constraints to avoid counterfactuals. Therefore, we need to rigorously consider the logical constraints between each event pair during prediction. To better measure the capability of LLMs on the ERE task, we formulate the logical consistency metric. 

\subsection{Logical Consistency Between Event Relations}
\label{subsec:lc}
Logical consistency plays a crucial role in accurate event relation prediction. In this paper, we consider a comprehensive set including 11 logical constraints applicable to all possible relations between two events, which are derived from realistic data and are detailed in Appendix~\ref{app:lc_of_two}. 
To quantify LLMs' adherence to these constraints, we introduce a metric called \textit{Logical Inconsistency} ($\mathbf{LI}$). This metric is calculated as the proportion of conflicts (i.e., the answers that conflict with the known logical constraints) to the total possible relation combinations (i.e., all combinations between any two relation types).  

To better illustrate the computation of $\mathbf{LI}$, here we introduce an example~(as shown in Figure~\ref{fig:example}): if an LLM outputs the relations between two events as ``NO\_COREFERENCE, SIMULTANEOUS, CAUSE, NO\_SUBEVENT''. Among these, ``SIMULTANEOUS'' and ``CAUSE'' are identified as conflicting with each other based on the logical constraints we have defined, creating an inconsistency. 
Considering there are four relation types to assess for each event pair, the total number of relation combinations is determined by the formula: $C_{4}^{2} = 6$. Thus in this example, with one identified conflict, $\mathbf{LI}$ is computed as  $1/6$ (or approximately 16.7\%). Based on the logical constraints, an algorithm can be designed to automatically detect conflicts and calculate the value of $\mathbf{LI}$.
Intuitively, the smaller the value of $\mathbf{LI}$ is, the more coherent and reasonable answer that LLM can produce.

\section{Unveiling LLMs in Logical Reasoning}
\label{sec:pilot_study}
Considering the rigorous logical reasoning required by ERE tasks, in this section, we conduct a pilot study to investigate how current LLMs exhibit reasoning tasks and how logic benefits LLMs.

\begin{figure}
\centering  
\includegraphics[width=0.48\textwidth]{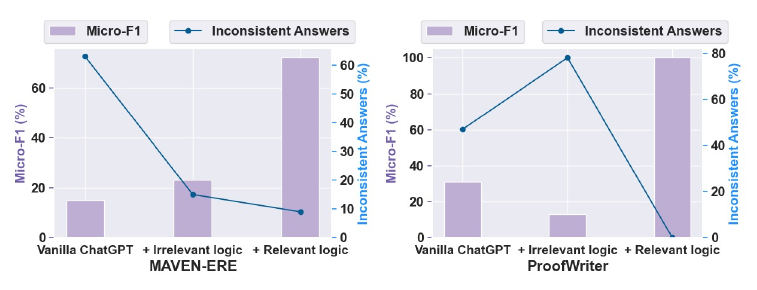}
\caption{Performance of ChatGPT in the pilot study.}
\label{fig:lce}
\end{figure}

\subsection{\textbf{Data Source}} We conduct a manual evaluation on MAVEN-ERE~\cite{wang2022maven} and ProofWriter~\cite{tafjord2020proofwriter}. MAVEN-ERE is a unified large-scale dataset for the ERE task, which needs to identify four types of relations. ProofWriter is a commonly used dataset for deductive reasoning, where each example is a pair of (problem, goal) and the label is selected from \{Proved, Disproved, Unknown\}.
To employ our investigation, we randomly choose 100 samples (50 from MAVEN-ERE and 50 from ProofWriter).
\subsection{\textbf{Experimental Setup}}
Our experiments are conducted in a zero-shot fashion.
Given a task input $(X)$, we also write a prompt $(T)$ describing the task, and let LLM generate output $(Y)$ by answering the given query.
We also add \emph{“Let’s think step by step”} before each answer for prediction generation, which is a simple but effective trick to improve zero-shot reasoning for LLMs~\cite{kojima2022large}. 
We adopt ChatGPT as the backbone and manually check its generated rationales under the following three settings:
\begin{itemize}[leftmargin=*]
    \item Vanilla LLM (i.e., ChatGPT) without any additional information;
    \item LLM (i.e., ChatGPT) plus the most relevant (i.e., ground truth) logic;
    \item LLM (i.e., ChatGPT) plus irrelevant logical constraints.
\end{itemize}
The latter two use a multi-turn conversational way based on the initial prediction from LLMs, so as to leverage LLM's interaction ability. The process of determining constraints for each way and the corresponding prompt examples can be found in Appendix~\ref{app:prompt_pilot}.
\begin{figure}
\centering  
\includegraphics[width=0.48\textwidth]{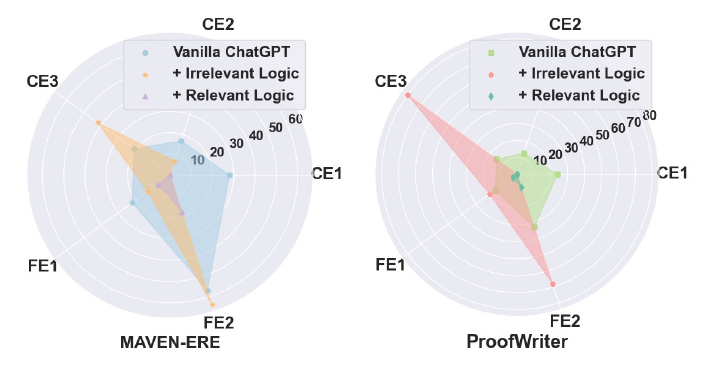}
\caption{Error analysis of ChatGPT in the pilot study by human evaluation. CE and FE denote incorrectness and unfaithfulness errors, respectively.}
\label{fig:error}
\end{figure}
\subsection{\textbf{Analysis}} 
As shown in Figure~\ref{fig:lce}, we visualize the micro-F1 values and the proportion of logically inconsistent answers generated by ChatGPT. We find that no matter whether on MAVEN-ERE or ProofWriter, Vanilla ChatGPT always achieves a bad result with low micro-F1 performance and high inconsistency values (e.g., 15\% micro-F1 and 63\% inconsistent answers on MAVEN-ERE), which indicates the deficiencies of LLM in solving complex reasoning tasks.
To investigate this issue in depth, we conduct analyses from the following two aspects.

\paragraph{\textbf{What is the Relation Between Logical Consistency and Model Performance?}}
From Figure~\ref{fig:lce}, we find that:
1) The model directly receives significant improvements on both MAVEN-ERE and ProofWriter when adding relevant logic; 
2) When adding some irrelevant logic, the results show some fluctuations (exaltation in MAVEN-ERE and degeneration in ProofWriter). That means directly adding logic without any constraints will bring some uncertainty; 
3) Typically, a higher logical inconsistency corresponds to a poorer micro-F1. However, rectifying logical inconsistency does not necessarily lead to the same degree of increase in micro-F1.
Generally, an intuitive observation is that incorporating relevant logic into the LLM instruction will be very helpful in solving reasoning tasks. Therefore, the challenges are how to obtain these relevant logic and how to utilize them for LLMs.

\paragraph{\textbf{What Types of Errors Does LLM Usually Make?}}
To delve into a deep understanding of the failures that vanilla LLM encounters in logical reasoning, we also conduct a detailed error analysis. Here, we divide the error types into two aspects: 
1) \emph{Incorrectness to the Constraint} (CE): whether the rationale generated by LLM is wrong (CE1), incomplete~(CE2), or redundant (CE3) compared with the true logical constraints.
2) \emph{Unfaithfulness to the Reasoning Process} (FE): where LLM does not correctly use the constraints. We define two types of errors upon FE, i.e., 
{i)} Wrong start, LLM begins with an irrelevant fact or focuses on an improper perspective for the correct answer (FE1).
{ii)} Wrong process, LLM starts from a proper point, but makes mistakes during the reasoning process (FE2).
Annotators are asked to review 100 predictions generated by ChatGPT and mark the error types. 
Results in Figure~\ref{fig:error} show that:
1) The quality of constraints produced by the vanilla ChatGPT is not high enough, which limits its subsequent reasoning ability.
2) Incorporating relevant logical constraints could guarantee the correctness of constraints and thus greatly improve the generation quality of ChatGPT in faithfulness.

\section{Teaching LLMs to Predict Event Relation Logic}

\begin{figure*}
\centering 
\includegraphics[width=0.9\textwidth]{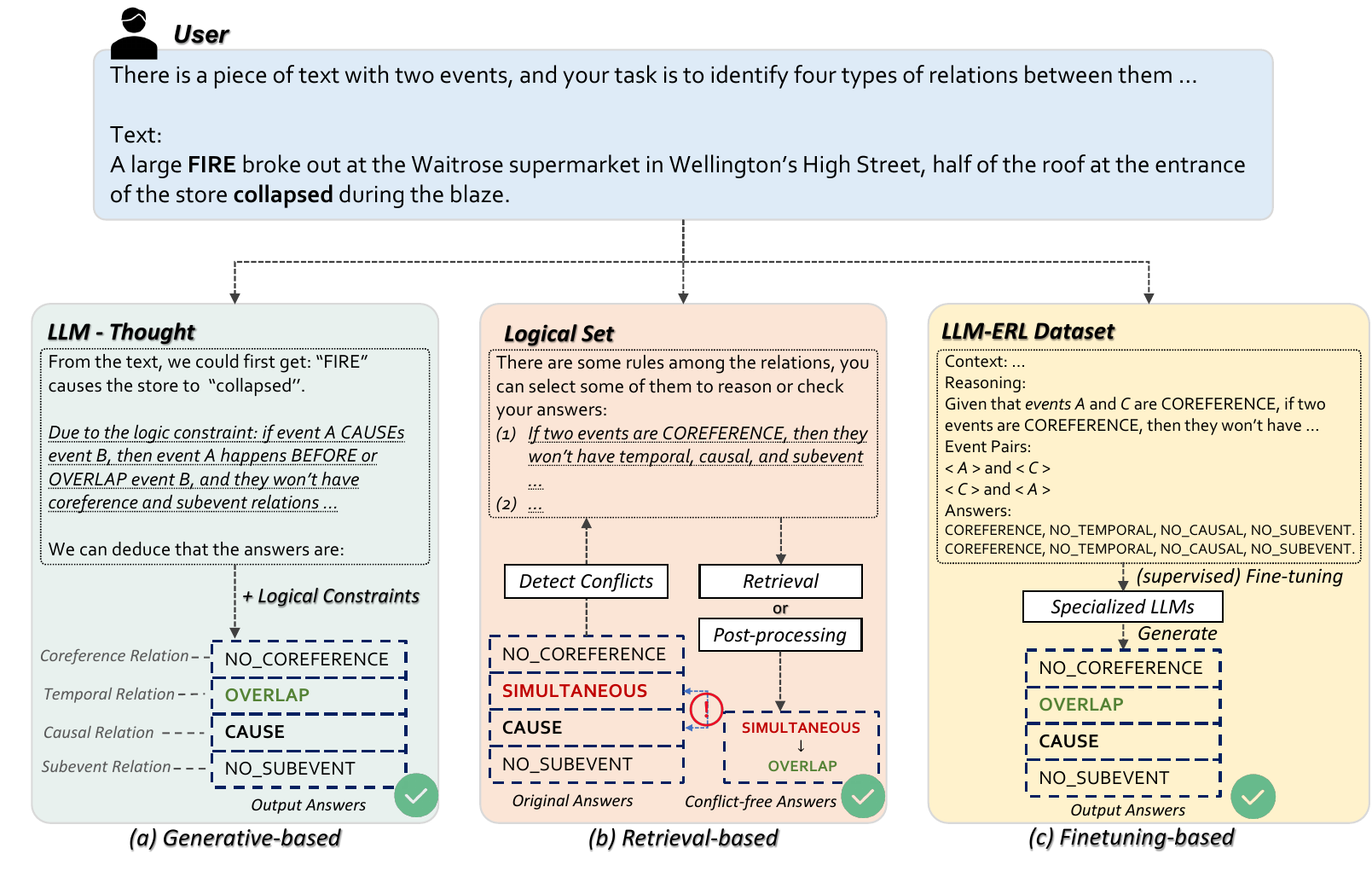} 
\caption{Incorporate logical constraints into LLMs by using generative, retrieval, and finetuning-based approaches. The dashed boxes indicate answers outputted by LLMs, and the underlined texts indicate the logical constraints.} 
\label{fig:prompts} 
\vspace{-4mm}
\end{figure*}
From the above analysis, the main reason for the failure of LLMs stems from their lack of logical reasoning abilities. 
In this section, we expect to explore how to augment LLMs with the capability to comprehend and apply event relation logic. Specifically, we first introduce the instruction-following technique used in Section~\ref{subsec:icl} and then propose three different approaches to instruct LLMs to generate answers with better logical consistency, including generative-based, retrieval-based, and finetuning-based approaches (Section \ref{subsec:generative} to \ref{subsec:pretraining}). We illustrate these three approaches in Figure~\ref{fig:prompts}.
\subsection{In-Context Learning for LLMs}
\label{subsec:icl}
We deploy LLMs for event relation logical prediction via in-context learning~(ICL, \cite{brown2020language, ouyang2022training}). Given a task input $(X)$, we write a prompt $(T)$ describing the task, then further provide several demonstrations $D=\left\{D_i\right\}_{i=1}^{|D|}$, where $D_i=\left(X_i, Y_i\right)$ are used for few-shot learning. Then, the LLM generates the output $(Y)$ by completing the prompt $(Y=\mathcal{M}(T, D, X))$, where $\mathcal{M}$ denotes the LLM. In such a setting, the LLM can follow the structure of the provided demonstrations to output answers in the expected format for subsequent automatic evaluation. 
Additionally, the whole process does not require any gradient update, allowing LLMs to generate predictions without massive training data.

\subsection{Generative-based Approaches}
\label{subsec:generative}
Generative-based approaches involve letting LLMs generate logic by using a form of few-shot ICL. Here, we study three variants:
\paragraph{Vanilla ICL:} which utilizes the common prompts consisting of the task description, the demonstration, and the input case. 

\paragraph{Vanilla CoT:} which first bootstraps rationales by using chain-of-thought as intermediate reasoning steps following the style of the given demonstration, then output answers. Rationales here do not involve the content of logical constraints. 

\paragraph{CoT with self-generated logical constraints:} which teaches LLMs to generate and utilize logical constraints based on CoT~(shown in Figure~\ref{fig:prompts} (a)). Specifically, it first extracts the obvious relations/facts and generates relevant logical constraints accordingly. LLMs are then prompted to infer the remaining relations and facts using these constraints along with the known information. An example prompt is provided in Appendix~\ref{app:prompt_lc}.

\subsection{Retrieval-based Approaches}
\label{subsec:retrieval}

Although generative-based approaches enable models to automatically generate and utilize logic, the generated rationales may be uncertain and inaccurate. Therefore, we also provide retrieval-based approaches, which aim to obtain relevant logic from our predefined logical set and add it to LLM instruction~(shown in Figure~\ref{fig:prompts} (b)). Specifically, we take all the constraints defined in Section~\ref{subsec:lc} as the retrieval set, and our solutions include:
\paragraph{with all logical constraints:} which directly adds all the text of logical constraints in the set. 
\paragraph{with retrieved logical constraints:} which means that we first detect logically inconsistent answers based on the prediction of LLMs, and then retrieve the corresponding information if we find any conflicts. Finally, we add the retrieved text to the LLM instruction and let LLMs regenerate the answers. Details can be found in Appendix~\ref{app:retrieval}. 
\paragraph{with post-processing:} which first obtains the answers of LLMs, then automatically generates some logically consistent candidates according to the known constraints, and randomly selects one of them as the final answer. This approach ensures that there are no logical conflicts ($\mathbf{LI} = 0\%$). Details can be found in Appendix~\ref{app:post-processing}.

\subsection{Finetuning-based Approach}
\label{subsec:pretraining}
Although the retrieval-based approach guarantees the correctness of logical constraints, it still needs to interact with an external logical set constantly. Therefore, we provide a finetuning-based approach to embed the logical constraints into LLMs themselves. Specifically, we first construct a high-order event relation logical prediction dataset \texttt{LLM-ERL}, then fine-tune specialized models on it, and finally use the fine-tuned models to conduct prediction. 

To construct \texttt{LLM-ERL}, we initiate with a foundational set of logical constraints for relations between two events that have been defined in Section \ref{subsec:lc}, and expand it to include additional constraints for high-order relations among three or more events based on \emph{transitive dependency}~\citep{allen1983maintaining, gerevini1995efficient}, i.e., one event may affect another through an intermediate event.  
The full transitivity rules are detailed in Appendix~\ref{app:lc_of_three} (Table~\ref{tab:lc_of_three}).

\paragraph{Dataset Construction}
Once obtaining the constraint set, the process of dataset construction becomes inferring new relations within a sequence of events based on any given relations. From there, we combine an initial relation with any other given relation to form a multi-hop query. This query aims to predict the logical outcome of a complex event interaction that spans multiple steps, leveraging the established logical constraints as a guide. For instance, if we have an initial relation ``BEFORE($A$, $B$)'', and we combine this with another two relations such as ``SIMULTANEOUS($B$, $C$)'' and  ``OVERLAP($C$, $D$)'', we are faced with a 3-hop query that seeks to deduce the relation between event $A$ and event $D$. Given the logical constraints, such as the transitivity rule that combines ``BEFORE'' and ``SIMULTANEOUS'' relations to infer new relations, we can deduce a logical outcome ``BEFORE($A$, $D$)''.
The corresponding pseudo-code can be found in Appendix~\ref{app:pseudo_code}.

The process of deducing the answer to these multi-hop queries is automated by employing logic programming~\citep{lloyd2012foundations, frederiksen2008applying}, specifically using forward- and backward-chaining methods in Prolog~\citep{clocksin2003programming}. This allows for the automatic inference of new relations based on the established set of logical constraints and the known relations among events. The outcome of this process can not only serve as the benchmark for evaluating or enhancing the reasoning capabilities of LLMs, but also act as a versatile platform for validating combinations of event relations across any number of hops.

\paragraph{Fine-tuning on \texttt{LLM-ERL}}
To fine-tune LLMs on \texttt{LLM-ERL}, 
we use the generated 2 to 5-hop reasoning data. We do not adopt longer hop data here considering the computation complexity and the length limitation of LLMs. We translate the symbolic representations of event relations into natural language descriptions to formulate queries,
aligning with the ERE task setup. This process resulted in a total of 6,776 instances. The dataset statistics are in Appendix~\ref{app:stat} and an illustrative example of such a prompt is depicted in Figure~\ref{fig:prompts} (c).
These queries not only promote LLMs' understanding of the logical constraints governing event sequences but also enhance their ability to apply these constraints in predicting the relations among events that are not explicitly given. Finally, we could conduct inference with the fine-tuned LLMs.

\section{Experiments}
\subsection{Experimental Setup}
\paragraph{Compared Models}
We choose several limited-access LLMs (\textit{gpt-3.5-turbo}, \textit{text-davinci-003}, and \textit{gpt-4}), and open-source LLMs (Vicuna-13B (v1.3)~\cite{vicuna2023} and Llama2-13B~\cite{touvron2023llama}) as the main experimental LLMs for evaluation. We also provide two fine-tuning RoBERTa-large~\cite{liu2019roberta} baselines (one-shot and fully fine-tuned) for comparison, the fine-tuning details can be found in Appendix~\ref{app:roberta}.

\paragraph{Dataset Construction}
Our main experiments are evaluated on two ERE datasets, MAVEN-ERE~\citep{wang2020joint} and Causal-TimeBank~\citep{mirza2014annotating}. All experiments are conducted in a one-shot fashion.
Further details can be found in Appendix~\ref{app:dataset}. 

\paragraph{\textbf{Fine-tuning Details}}
For the finetuning-based approach, we adopt Vicuna-13B (v1.3) and Llama2-13B as the base models and employ the LoRA~\citep{hu2022lora} technique. During fine-tuning, only LoRA parameters are optimized. The fined-tuned models are named Vicuna-FT and Llama2-FT, respectively. Further details can be found in Appendix~\ref{app:pretraining}.

\paragraph{Evaluation Metrics}
We adopt the averaged micro-F1 score as the evaluation metric and also report the logical inconsistency metric $\mathbf{LI}$ (defined in Section~\ref{subsec:lc}) on ERE datasets. The reported value is averaged by the results of three runs to reduce random fluctuation.

\subsection{Main Results}
\begin{table*}[!t]
    \renewcommand
    \arraystretch{1.0}
    \centering
    \small
    \setlength{\tabcolsep}{8pt}
        \begin{tabular}{ll|cc|cc}
        \toprule
        \multicolumn{2}{l|}{\multirow{2}*\textbf{Model}}
         & \multicolumn{2}{c|}{\textbf{MAVEN-ERE}} & \multicolumn{2}{c}{\textbf{Causal-TimeBank}}   \\ 
         \cmidrule(lr){3-4}\cmidrule(l){5-6} 
          & &Micro-F1 (\%)  & $\mathbf{LI}$ (\%) $\downarrow$ &Micro-F1 (\%) & $\mathbf{LI}$ (\%) $\downarrow$ \\ \midrule
        \multicolumn{2}{l|}{\textbf{RoBERTa-Large (fully fine-tuned)}} &56.8 &6.4 &22.2 &36.2  \\
        \multicolumn{2}{l|}{\textbf{RoBERTa-Large (one-shot)}} &17.4 &54.8 &- &-  \\
        \midrule
        \multirow{6}{*}{\bf{Turbo}} & \cellcolor{generative} vanilla ICL &18.0  &53.3  &19.0 &54.0  \\
        & \cellcolor{generative} vanilla CoT &18.8  &49.3  &17.0  &30.3  \\
        & \cellcolor{generative} CoT w. logical constraints &\textbf{25.3}  &{37.9}  &\textbf{27.0}  &{12.8}   \\
         \cmidrule(lr){2-2}\cmidrule(lr){3-4}\cmidrule(l){5-6}
        & \cellcolor{retrieval} w. all logical constraints &{20.8} &30.9  &{20.0} &36.8 \\
        & \cellcolor{retrieval} w. retrieved logical constraints  &{22.3}  &{30.2} &{22.0} &{11.3} \\
        & \cellcolor{retrieval} w. post-processing &14.0  &\textbf{0}  &15.0 &\textbf{0} \\
         \midrule
        \multirow{6}{*}{\bf{Davinci}} 
        & \cellcolor{generative} vanilla ICL &21.6  &49.1  &18.0 &58.8 \\
        & \cellcolor{generative} vanilla CoT &20.5  &60.5  &21.0 &64.7   \\
        & \cellcolor{generative} CoT w. logical constraints &{24.8}  &{5.5}  &{23.0} & {39.2}\\
         \cmidrule(lr){2-2}\cmidrule(lr){3-4}\cmidrule(l){5-6}
        & \cellcolor{retrieval} w. all logical constraints &{27.0}  &25.6  &\textbf{31.0} & {21.8}\\
        & \cellcolor{retrieval} w. retrieved logical constraints  &\textbf{27.8}  &30.8  &{22.0} &40.5 \\
        & \cellcolor{retrieval} w. post-processing &14.8  &\textbf{0}  &19.0 &\textbf{0} \\
        \midrule
        \multirow{6}{*}{\bf{GPT-4}} 
        &\cellcolor{generative} vanilla ICL &29.3  &50.7  &22.5 &30.5\\
        & \cellcolor{generative} vanilla CoT &30.3  &36.7  &23.0 &35.0   \\
        & \cellcolor{generative} CoT w. logical constraints &{32.3}  &{13.7}  &{24.5} & {24.0} \\
         \cmidrule(lr){2-2}\cmidrule(lr){3-4}\cmidrule(l){5-6}
         & \cellcolor{retrieval} w. all logical constraints &\textbf{37.3}  &{8.3}  &\textbf{26.0} & {20.0}\\
        & \cellcolor{retrieval} w. retrieved logical constraints  &{33.5}  &28.8  &{24.0} &{13.5} \\
        & \cellcolor{retrieval} w. post-processing &17.0  &\textbf{0}  &19.0 &\textbf{0} \\
        \midrule
        \multirow{6}{*}{\bf{Vicuna}} 
        &\cellcolor{generative} vanilla ICL &13.8  &25.4  &4.5 &84.1 \\
        & \cellcolor{generative} vanilla CoT &11.6  &47.4  &6.0 &57.6   \\
        & \cellcolor{generative} CoT w. logical constraints &{14.9}  &{21.7}  &{8.0} &{33.1}  \\
         \cmidrule(lr){2-2}\cmidrule(lr){3-4}\cmidrule(l){5-6}
        & \cellcolor{retrieval} w. all logical constraints &{15.2}  &37.6  &\textbf{11.0} &{23.5} \\
        & \cellcolor{retrieval} w. retrieved logical constraints  &\textbf{15.7}  &33.2  &{10.0} &26.7 \\
        & \cellcolor{retrieval} w. post-processing &9.8  &\textbf{0}  &9.0 &\textbf{0} \\
        \midrule
        \multirow{6}{*}{\bf{Llama2}} 
        &\cellcolor{generative} vanilla ICL &17.0  &54.6  &{11.5} &26.7\\
        & \cellcolor{generative} vanilla CoT &17.8  &58.4  &10.5 &33.6   \\
        & \cellcolor{generative} CoT w. logical constraints &\textbf{21.5}  &{18.9}  &\textbf{13.0} &{18.1}  \\
         \cmidrule(lr){2-2}\cmidrule(lr){3-4}\cmidrule(l){5-6}
        & \cellcolor{retrieval} w. all logical constraints &{19.5}  &34.6  &{10.0} &{23.5} \\
        & \cellcolor{retrieval} w. retrieved logical constraints  &{18.3}  &38.2  &{9.5} &26.7 \\
        & \cellcolor{retrieval} w. post-processing &12.0  &\textbf{0}  &{9.5} &\textbf{0} \\
        \midrule
        \multirow{6}{*}{\bf{Vicuna-FT}}
        &\cellcolor{pretraining} vanilla ICL &15.3   &21.2   &8.0  &35.5  \\
        & \cellcolor{pretraining} vanilla CoT &15.8  &17.8  &7.5  &52.5   \\
        & \cellcolor{pretraining} CoT w. logical constraints &\textbf{18.0}  &{6.0 }  &8.5 &2.0   \\
        \cmidrule(lr){2-2}\cmidrule(lr){3-4}\cmidrule(l){5-6}
        & \cellcolor{pretraining} w. all logical constraints &{16.3}  &8.7  &\textbf{12.1} &\textbf{0} \\
        & \cellcolor{pretraining} w. retrieved logical constraints  &{16.1}  &19.0   &{10.7} &9.5 \\
        & \cellcolor{pretraining} w. post-processing &11.0   &\textbf{0}  &8.0  &\textbf{0} \\
        \midrule
       \multirow{6}{*}{\bf{Llama2-FT}} 
        &\cellcolor{pretraining} vanilla ICL &19.0  &45.8  &{12.0} &22.7 \\
        & \cellcolor{pretraining} vanilla CoT &22.1  &42.9  &11.5  &3.0   \\
        & \cellcolor{pretraining} CoT w. logical constraints &\textbf{26.4 }  &{15.7}  &\textbf{13.3 } &{13.0}  \\
        \cmidrule(lr){2-2}\cmidrule(lr){3-4}\cmidrule(l){5-6}
        & \cellcolor{pretraining} w. all logical constraints &{20.2 }  &28.7   &{12.0 } &{23.0 } \\
        & \cellcolor{pretraining} w. retrieved logical constraints  &{18.7 }  &34.2  &{11.0 } &19.4  \\
        & \cellcolor{pretraining} w. post-processing &11.0  &\textbf{0 }  &11.0  &\textbf{0} \\
        \bottomrule
        \end{tabular}
       \caption{ \label{tab:lce-all} Proprietary LLMs (\textit{gpt-3.5-turbo}, \textit{text-davinci-003}, and \textit{gpt-4}), Vicuna-13B, Llama2-13B's performance on MAVEN-ERE and Causal-TimeBank. ``PT'' denotes after fine-tuning on \texttt{LLM-ERL}. For each dataset, the best result of each LLM is in \textbf{bold}. RoBERTa-Large (one-shot) fails to output any correct answers on Causal-TimeBank. The highlighted colors denote \sethlcolor{generative}\hl{generative-based}, \sethlcolor{retrieval}\hl{retrieval-based}, and \sethlcolor{pretraining}\hl{finetuning-based} approaches, respectively.} 
\end{table*}
From Table~\ref{tab:lce-all}, We could observe that: 
\paragraph{Generative-based Approaches}

\noindent1) Compared with a smaller language model RoBERTa-large, the generalization ability of vanilla LLMs under the one-shot setting is remarkable, but there is still a gap with the fully-finetuned baseline.

2) Directly using CoT to infer logic does not help much for ERE tasks, a possible reason is that the inherent issues of LLMs may cause them to fail in generating precise rationales (i.e., a high ratio of logical inconsistency). 

3) When using generative-based approaches to encourage LLMs to produce logical constraints in the reasoning process, LLMs can significantly improve their performance on ERE tasks (e.g., 7.3\% F1 performance gains from 18.0\% to 25.3\% of \textit{gpt-3.5-turbo} on MAVEN-ERE). We give a case study for the generative-based approach in Appendix~\ref{app:case}, which shows how LLMs perform when generating logical constraints by themselves.

\paragraph{Retrieval-based Approaches}

\noindent1) When using retrieval-based approaches to obtain logic constraints and incorporate them into LLM instruction, the logical inconsistency of LLMs' answers is greatly reduced and the overall performance is further improved (e.g., 6.2\% F1 performance gains from 21.6\% to 27.8\%, and 18.3\% $\mathbf{LI}$ decrease\ from 49.1\% to 30.8\% of \textit{text-davinci-003} on the MAVEN-ERE dataset).

2) Among all the limited-access models, we find that only \textit{gpt-4} perform better under the \emph{``w. all logical constraints''} setting compared with the \emph{``w. retrieved logical constraints''} setting. We hypothesize that this is due to the superior language understanding and retrieval capabilities of  \textit{gpt-4}, enabling it to identify some useful logical constraints to derive the answers accurately. In contrast, earlier models may struggle to filter out irrelevant information and therefore still require our assistance in retrieval to screen the necessary information.

3) Although the post-processing baseline guarantees the absence of logical conflicts (resulting in $\mathbf{LI}$ of 0\%), it may severely affect the quality of the whole generation. On one hand, the semantics of the post-processing answer may be far from the ground truth due to the random selection. On the other hand, the size of the candidate set for each case will also affect the performance. 
It may also need more operations at the post-processing stage, which we leave as future work.
We also conduct ablation studies on the number of demonstration samples and iterative retrievals in Section~\ref{subsec:ablation}.

\paragraph{Finetuning-based Approach}

\noindent1) Once fine-tuned on \texttt{LLM-ERL}, the performance of Llama2-FT and Vicuna-FT improves greatly compared with vanilla Llama2 and Vicuna, especially on the baselines without logical constraints.

2) The performance of Llama2-FT (i.e., 26.4\% F1 score on MAVEN-ERE) could even surpass that of some greater LLMs (e.g., vanilla \textit{gpt-3.5-turbo}, 25.3\%), which further validates the importance of teaching LLM with event relation logic in solving ERE tasks. We also conduct a case study comparing the output answers of Llama2 and Llama2-FT in Appendix~\ref{app:llama-case}.

\subsection{Ablation Study}
\label{subsec:ablation}
\begin{figure}
\centering  
\includegraphics[width=0.48\textwidth]{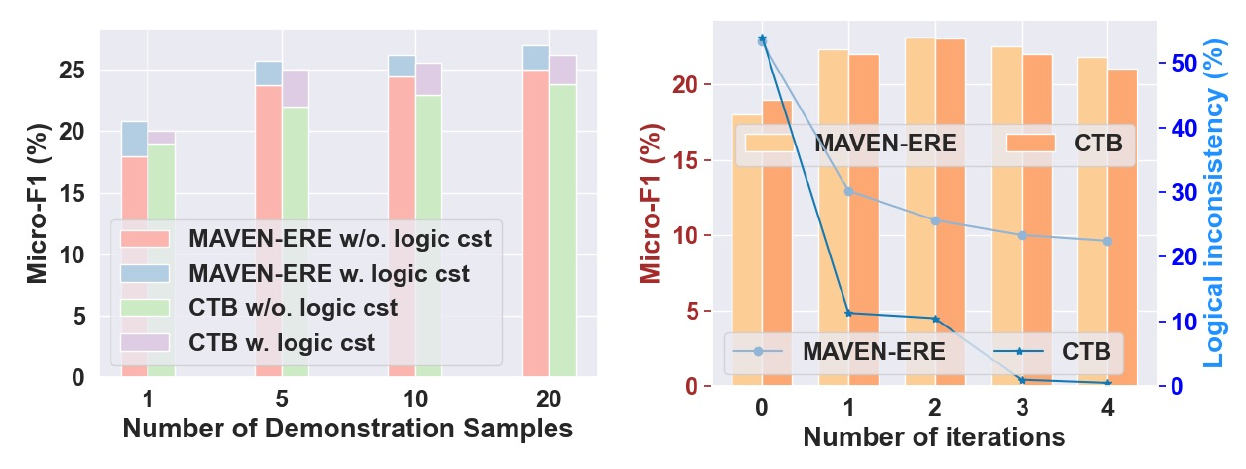}
\caption{Ablation Study of ChatGPT for demonstrations and iterative retrieval, where ``logic cst'' denotes the event relation logical constraints.}
\label{fig:ablation}
\end{figure}

We conduct an ablation study using ChatGPT~(\textit{gpt-3.5-turbo}) in this section.
\paragraph{Demonstrations}
Following previous experiences~\citep{brown2020language}, we also append demonstrations into the prompt to investigate how logical constraints will affect when combined with different numbers of demonstrations. Here, we select different numbers of demonstration samples $K$ from $\{1, 5, 10, 20\}$. The experiments are tested on the \emph{``w. all logical constraints''} settings, and we choose the \emph{``vanilla ICL''} baseline for comparison. 
From Figure~\ref{fig:ablation} (left), we can observe that: 
1)  When the number of demonstrations increases from 1 to 5, there is an evident performance improvement, but the subsequent improvements are limited when continue to increase the number of demonstrations (e.g., $\geq$ 10); 
2) Adding logical constraints into LLM instructions can provide stable improvements, especially with more demonstrations; 
3) The performance of incorporating logical constraints with a smaller number of demonstrations can even surpass that of prompts with only a larger number of demonstrations~(e.g., the F1 performance of using 5 demonstrations on MAVEN-ERE w. logical constraints, 25.7\%, surpasses that of 10 demonstrations w/o. logical constraints, 24.5\%). This indicates that it is important to tell LLMs both ``What''~(demonstrations) and ``How''~(logical constraints).
Overall, these studies further confirm the merits of using event relation logic in solving ERE tasks.

\paragraph{Iterative Retrieval} 
Considering the outstanding ability of LLMs in interaction, we further explore whether we can introduce logical constraints into the multi-turn conversation (for the prompt design, please see Appendix~\ref{app:prompt_iterative}). Here, we adopt a retrieval-based approach to incorporate retrieved logical constraints iteratively and the results are shown in Figure~\ref{fig:ablation} (right). We find that the logical inconsistency of answers will gradually decrease with the increase of iterations, but the overall micro-F1 score seems relatively stable. We guess the main reason for this phenomenon is the overthinking of LLMs, as although it can bring more reasoning rationale, it possibly produces correct but more useless or abundant information when inferring multiple iterations. Overall, instructing LLM with logic is beneficial for conversation, but how to support longer information is still challenging.

\section{Related Work}
\subsection{Large Language Models (LLMs)}
We are fortunate to witness the surging development of Large Language Models (LLMs~\cite{brown2020language, ouyang2022training, chowdhery2022palm, chung2022scaling}), and a series of work aiming to leverage the reasoning abilities of LLMs such as chain-of-thought prompting~\cite{wei2022chain, 
 kojima2022large, zhang2022automatic}, self verification~\cite{wang2022self, jung-etal-2022-maieutic}, self learning~\cite{zelikman2022star, huang2022large}, etc. However, recent studies show LLMs still stumble in generating hallucination and logic inconsistency~\cite{golovneva2022roscoe, jang2023consistency, bang2023multitask, liu2023evaluating, jiao2023logicllm}. To solve such challenges, our work explores teaching LLMs logical reasoning through various approaches.
 
\subsection{Event Relation Extraction (ERE)}
Events play crucial roles in comprehending narratives, and understanding the complex relationships between events is essential to understanding the text~\cite{sundheim-1992-evaluating}. Thus ERE tasks are fundamental information extraction (IE) tasks and support various downstream applications~\cite{chaturvedi2017story, zhang2020transomcs}. Extensive studies have been carried out on ERE tasks, including different kinds of relations such as coreference relations~\cite{lu2021conundrums, lu2022end}, temporal relations~\cite{ning2018joint, wang2020joint,han-etal-2019-deep, zhou2021clinical}, causal relations~\cite{caselli2017event, chen2022ergo, chen2023cheer}, and subevent relations~\cite{aldawsari2019detecting, wang2021learning}.

There also have been some recent explorations on how to leverage the power of LLMs on event-related information extraction tasks~\cite{wang2022code4struct, 2023_chatgpt_ee, ma2023large, qiu2023large, yuan2024back}. To the best of our knowledge, however, our work is the first to 1) design elaborate experiments to evaluate the performance of LLMs on the ERE task, including coreference, temporal, causal, and subevent relations, 2) delve into the high-order logical constraints between these event relations, and (3) analyze the logical reasoning abilities of LLMs using ERE as an intermediate task.

\section{Conclusion}
In this paper, we conduct a detailed investigation on how to enhance LLMs with event relation logic. Specifically, we first investigate the existing issues of current LLMs in event relation logical prediction. Then, we study multiple strategies to obtain and utilize logic for LLMs, including generative-based, retrieval-based, and finetuning-based approaches. Based on
our approaches, we also contribute a synthesized dataset (\texttt{LLM-ERL}) involving multi-hop reasoning for evaluation and fine-tuning.
We show that LLMs are not logically consistent reasoners, but their performance could be improved if we explicitly teach them the logical constraints. Comprehensive quantitative and qualitative analyses have been conducted to further provide insights.

\section*{Limitations}
Although we have explored a series of approaches in detail to enhance LLMs to generate more logically consistent answers and greatly improve their performance, we find that there is still a certain gap between this and the ideal situation (i.e., incorporating the most relevant logical constraints in Section~\ref{sec:pilot_study}). In view of the LLMs' potential to understand logical constraints and make more rigorous reasoning, we believe that further exploration of how to make better use of logical constraints will help us understand the reasoning ability of LLMs, and we will take this as our future work.

\section*{Acknowledgments}
We thank all the anonymous reviewers for their valuable feedback throughout the review process.
This work is also supported by Ucap Cloud.
\bibliography{custom}
\clearpage
\appendix
\section{Understanding Event Relations}
\label{app:temp}
\begin{figure}
\centering  
\includegraphics[width=0.48\textwidth]{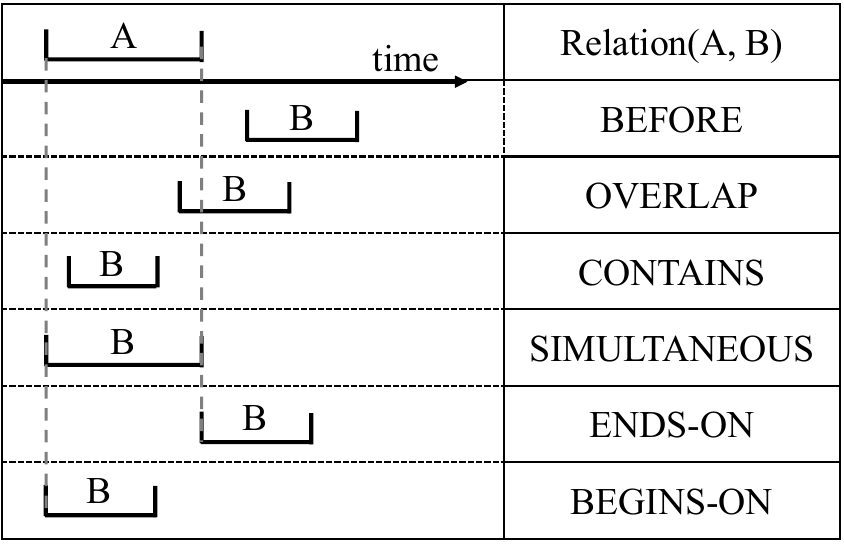}
\caption{Interpretations of the temporal relation between two events A and B. Brackets represent time intervals along the time axis.}
\label{fig:temp_rels}
\end{figure}

There are four kinds of widely-used event relations: coreference, temporal, causal, and subevent relations~\cite{o2016richer, wang2022maven}. 
\begin{enumerate}[leftmargin=*]
    \item \emph{Coreference relations} between events occur when multiple event mentions in a text refer to the same underlying event. We call these event mentions \textit{cluster}. 
    \item \emph{Temporal relations} refer to the temporal ordering of events based on their occurrence in time. In this paper, we consider seven different types of temporal relations:
\begin{itemize}
    \item NO\_TEMPORAL: if there is no clear temporal relation between event $A$ and $B$.
    \item BEFORE: if event $A$ happened completely before event $B$.
    \item OVERLAP: if event $A$ has an overlap with event $B$. 
    \item CONTAINS: if event $A$'s time contains event $B$'s time. 
    \item SIMULTANEOUS: if events A and $B$ happen at the same time. \item ENDS-ON: if event $A$ ends when event $B$ starts. 
    \item  BEGINS-ON: if event $A$ and event $B$ start at the same time, but end at different times. 
\end{itemize}
In Figure~\ref{fig:temp_rels}, we list all the types of temporal relations and illustrate their distinctions on a unified timeline. Note that in our study, we adhere to a unidirectional perspective where the start time of event $A$ precedes that of event $B$. Consequently, our framework does not encompass symmetrical relationships, such as the inverse of ``AFTER'' being ``BEFORE''. To illustrate, if event $A$ is considered ``AFTER'' event 
$B$, this would correspond to event $B$ being ``BEFORE'' event $A$ in our defined context.

    \item \emph{Causal relations} refer to that one event (the cause) brings about or influences the occurrence of another event (the effect). They can be classified into two different types: \textit{CAUSE} relation where the tail event is inevitable given the head event, and \textit{PRECONDITION} where the tail event would not have happened if the head event had not happened. 
    \item \emph{Subevent relations} refer to that one event (the subevent) is a component or a smaller part of another event (the main event). Identifying and understanding subevent relations helps to reveal the underlying hierarchy and organizational structure of events in a given text.
\end{enumerate}

\paragraph{Event Relation Extraction}
Event Relation Extraction~(ERE) includes identifying coreference, temporal, causal, and subevent relations between every two events in the text. We formulate ERE as a multi-label classification problem, determining one label (relation) for each of these four relation types. For coreference relations, the labels $\in$\{NO\_COREFERENCE, COREFERENCE\}; for temporal relations, the labels $\in$ \{NO\_TEMPORAL, BEFORE, OVERLAP, CONTAINS, SIMULTANEOUS, ENDS-ON, BEGINS-ON\}; for causal relations, the labels $\in$ \{NO\_CAUSAL, PRECONDITION, CAUSE\}; for subevent relations, the labels $\in$ \{NO\_SUBEVENT, SUBEVENT\}.

\section{Logical Constraints Between Two Events}
\label{app:lc_of_two}
In Table~\ref{tab:lc_of_two}, we provide a comprehensive set of logical constraints for the relations between two events to assess their logical consistency. We also manually design description text for each constraint to let LLMs follow the prompt. As shown in Table~\ref{tab:lc_of_two_txt}, COREFERENCE($A$, $B$) $\rightarrow$ $\neg$TEMPORAL($A$, $B$), $\neg$CAUSAL($A$, $B$),  $\neg$SUBEVENT($A$, $B$) indicates that "if event $A$ and event $B$ have a coreference relation, they will not have temporal, causal, and subevent relations". 

\subsection{An Example of Detecting Conflicts and Retrieving Relevant Constraints}
\label{app:retrieval}
As described above, for the ERE task, we meticulously collect 11 logical constraints covering all relations between two events. These constraints serve as our benchmark to identify inconsistencies in the predictions made by LLMs. 

Let us consider an illustrative example. If LLM produces an answer such as ``NO\_COREFERENCE, SIMULTANEOUS, CAUSE, NO\_SUBEVENT'' (refer to Figure~\ref{fig:example} and Figure~\ref{fig:prompts}), we could detect the inconsistency between ``SIMULTANEOUS'' and ``CAUSE'', as shown in Table ~\ref{tab:lc_of_two}:
\begin{itemize}
    \item A ``SIMULTANEOUS''  relation implies a ``NO\_CAUSAL'' (¬CAUSAL) relation.
    \item Conversely, a ``CAUSE'' relation suggests the presence of either a ``BEFORE'' or an ``OVERLAP'' relation.
\end{itemize}
Given this, ``SIMULTANEOUS'' and ``CAUSE'' are inherently contradictory, and they cannot coexist in a consistent prediction. To rectify this, we retrieve the associated textual description from Table~\ref{tab:lc_of_two_txt}. Specifically, the statements ``\emph{If event $A$ CAUSEs event $B$, then event $A$ happens BEFORE or OVERLAP event $B$ ...}'' and ``\emph{If event $A$ and event $B$ happen SIMULTANEOUSly,
then they won’t have coreference, causal, and subevent relations ...}''  are integrated into the LLM's instruction. 

\begin{table*}[htbp!]
    \renewcommand
    \arraystretch{1.0}
    \centering
        \small
    \setlength{\tabcolsep}{6pt}
    \begin{tabular}{l|c|c}
    \toprule
         {\textbf{If Relation($A$, $B$)}}  & \textbf{Then Relation ($A$, $B$)} & \textbf{Then Relation ($B$, $A$)} \\ \midrule
         COREFERENCE & $\neg$TEMPORAL, $\neg$CAUSAL,  $\neg$SUBEVENT &COREFERENCE \\ 
         $\neg$TEMPORAL &$\neg$CAUSAL,  $\neg$SUBEVENT & / \\
         BEFORE &$\neg$COREFERENCE, $\neg$SUBEVENT &$\neg$TEMPORAL\\
         OVERLAP &$\neg$COREFERENCE, $\neg$SUBEVENT &$\neg$TEMPORAL\\ 
         CONTAINS &$\neg$COREFERENCE, $\neg$CAUSAL &$\neg$TEMPORAL\\ 
         SIMULTANEOUS & $\neg$COREFERENCE,  $\neg$CAUSAL,  $\neg$SUBEVENT &SIMULTANEOUS \\ 
         ENDS-ON &$\neg$COREFERENCE, $\neg$CAUSAL, $\neg$SUBEVENT &$\neg$TEMPORAL\\
         BEGINS-ON &$\neg$COREFERENCE, $\neg$CAUSAL, $\neg$SUBEVENT &BEGINS-ON\\
         CAUSE &$\neg$COREFERENCE, BEFORE $\vee$ OVERLAP,  $\neg$SUBEVENT &$\neg$TEMPORAL\\ 
         PRECONDITION &$\neg$COREFERENCE, BEFORE $\vee$ OVERLAP, $\neg$SUBEVENT&$\neg$TEMPORAL\\ 
         SUBEVENT &$\neg$COREFERENCE, CONTAINS $\neg$CAUSAL& $\neg$TEMPORAL\\ 
    \bottomrule
    \end{tabular}
       \caption{\label{tab:lc_of_two} Logical Constraints of relations between two events, where $\neg$ denotes "NOT", $\vee$ denotes "OR".}
\end{table*}

\subsection{An Example of Post-processing}
\label{app:post-processing}
As shown in Figure~\ref{fig:prompts}, if LLMs predict the relations between two events as ``NO\_COREFERENCE, SIMULTANEOUS, CAUSE, NO\_SUBEVENT'', we can detect that ``SIMULTANEOUS'' and ``CAUSE'' conflict according to the logical constraints. In order to eliminate conflicts, one relation can be fixed first, and then the other relation can be randomly decided by the candidates that do not conflict with the current relation. For example, when the fixed temporal relation is ``SIMULTANEOUS'', the causal relations can only be ``NO\_CAUSAL'', while when the fixed causal relation is ``CAUSE'', the temporal relation can be either ``BEFORE'' or ``OVERLAP''. We also add a negative option ``NO\_COREFERENCE, NO\_TEMPORAL, NO\_CAUSAL, NO\_SUBEVENT'' to the candidate set because it is possible that neither relation exits. Finally, we randomly select one option from:

 \begin{itemize}
   \item NO\_COREFERENCE, SIMULTANEOUS, NO\_CAUSAL,  NO\_SUBEVENT
    \item NO\_COREFERENCE, OVERLAP, CAUSE, NO\_SUBEVENT
    \item  NO\_COREFERENCE, BEFORE, CAUSE, NO\_SUBEVENT
    \item NO\_COREFERENCE, NO\_TEMPORAL, NO\_CAUSAL, NO\_SUBEVENT
\end{itemize}
as the ultimate answer,  thus ensuring that the results must be logically consistent (i.e., $\mathbf{LI} = 0$).

\section{Transitivity Rules Among Events}
\label{app:lc_of_three}
We provide a comprehensive set of 39 logical constraints for the transitivity rules among three events in Table~\ref{tab:lc_of_three}. We also manually design prompts for each constraint, as shown in Table~\ref{tab:lc_of_three_txt}.

\subsection{Pseudo Code of Logic Programming}
\label{app:pseudo_code}
Once obtaining 11 constraints between two events and 39 constraints among three events, we apply logic programming to automatically reason new event relations by inputting the known constraints and relations. The pseudo-code mentioned in the main text is shown in Algorithm~\ref{alg1}.

\section{Statistics of the Fine-tuning Dataset}
\label{app:stat}
As shown in Table \ref{tab:llm-lr}, we provide the statistics of the fine-tuning dataset originating from \texttt{LLM-ERL}.
\begin{table}[htbp!]
    \renewcommand
    \arraystretch{1.0}
    \centering
    \setlength{\tabcolsep}{10pt}
        \begin{tabular}{l|c}
        \toprule
        \textbf{Hop}
         & \textbf{\# Count} \\ 
        \midrule
        2 &39  \\
        3 &179 \\
        4 &945 \\
        5 &5613 \\
        \bottomrule
        \end{tabular}
       \caption{ \label{tab:llm-lr} Statistics of the fine-tuning dataset.}
\end{table}

\section{Dataset Construction}
\label{app:dataset}
\paragraph{MAVEN-ERE} contains 4,480 documents, 103,193 events coreference chains, 1,216,217 temporal relations, 57,992 causal relations, and 15,841 subevent relations, which is larger than existing datasets of all the ERE tasks by at least an order of magnitude~\cite{wang2022maven}. MAVEN-ERE has released the train and valid set, but does not release the ground-truth test set, so we randomly split its train set into train/valid sets with a ratio of 8:2, and then use its original valid set as the new test set.

\paragraph{Causal-TimeBank} contains 184 documents, 6,813 events, and  7,608 event pairs~\cite{mirza2014analysis}. Among them, 318 and 6,115 event pairs are annotated with causal and temporal relations, respectively. Due to Causal-TimeBank does not split train/valid/test sets, we randomly split it to train/valid/test sets with a ratio of 6:1:3. We do not evaluate coreference and subevent relations in Causal-TimeBank since there are no annotations for these two relation types.

For ERE tasks, We conduct sampling at the sentence level. The samples of the two events that do not have any relations will be excluded. Note that Causal-TimeBank inherently contains fewer event relations compared to MAVEN-ERE. After processing and dividing the data split, its test set comprises only 139 samples. Therefore, we randomly sample 500 examples from the test set of MAVEN-ERE and 100 examples from the test set of Causal-TimeBank as our testbed. 

\paragraph{ProofWriter} is a commonly used dataset for deductive reasoning~\citep{tafjord2020proofwriter}. We use the OWA subset of it, which is divided into five parts, each part requiring 0, 1, 2, 3, and 5 hops of reasoning, respectively. We evaluate the hardest 5-hop subset. To reduce the computation cost, we randomly sample 200 examples in the test set and ensure a balanced label distribution. 

\paragraph{FOLIO} is a challenging expert-written dataset for logical reasoning~\citep{tafjord2020proofwriter}, whose questions require complex first-order logic reasoning to solve. We use its entire test set for evaluation, consisting of 204 examples.

\renewcommand{\algorithmicforall}{\textbf{for each}}
\begin{algorithm}[htbp!]
\caption{An Example of 3-hop Reasoning}
\label{alg1} 
\begin{algorithmic}

\STATE Initialize the knowledge base with facts and rules
\STATE \hspace{1em} Knowledge Base:
\STATE \hspace{2em} Fact: BEFORE($A$, $B$)
\STATE \hspace{2em} Fact: SIMULTANEOUS($B$, $C$)
\STATE \hspace{2em} Fact: OVERLAP($C$, $D$)
\vspace{1mm}
\STATE \hspace{2em} Rule: BEFORE $\leftarrow$ BEFORE $\wedge$ SIMULTANEOUS
\STATE \hspace{2em} Rule: OVERLAP $\leftarrow$ SIMULTANEOUS $\wedge$ OVERLAP 
\STATE \hspace{2em} Rule: BEFORE $\leftarrow$ BEFORE $\wedge$ OVERLAP
\vspace{2mm}
\STATE Initialize the logic engine with the query
\STATE \hspace{1em}Query: BEFORE($A$, $D$)?
\vspace{2mm}
\WHILE{obtain new facts}
    \FORALL{rule $r$ of the Knowledge Base}
        \IF{$r$'s premise is satisfied by the current known facts}
            \STATE \textbf{Add} $r$'s conclusion to the knowledge base
        \ENDIF
    \ENDFOR
\ENDWHILE
\vspace{2mm}
\STATE Query result: BEFORE($A$, $D$) is satisfied with BEFORE($A$, $C$) and OVERLAP($B$, $D$)

\end{algorithmic}
\end{algorithm}
\section{Training Details of RoBERTa-large On Two Tasks}
\label{app:roberta}
Our experiments include two settings. (1) fully fine-tuned: we fine-tune smaller language models (SLMs) with complete and abundant samples. This setting is for reference to see the performance limit of SLMs. (2) one-shot: we sample only one example for each label and construct a tiny training set. This setting is for direct comparison with our experiments on LLMs (similar training/demonstration sample number).

We implement vanilla fine-tuning approaches on two datasets and use RoBERTa-Large as backbones. We run each experiment on a single NVIDIA V100 GPU. We adopt the AdamW~\cite{loshchilov2017decoupled} optimizer with a linear scheduler and 0.1 warm-up steps. We set the weight-decay coefficient as 1e-5 and maximum gradient norms as 1.0. We set the batch size as 16 with 20 or 50 epochs. We set the maximum input length as 256 and the
learning rate as 2e-5.

\section{Implementation Details of Finetuning-based Approach}
\label{app:pretraining}
We set the rank of LoRA modules to be 64. Our model is optimized with a learning rate of 2e-4 and a linear
warm-up for the first 3\% steps. We clip the gradients of model parameters to a max norm of 0.3.
All the LoRA parameters are fine-tuned on an NVIDIA A100 GPU with 80GB memory.

\section{Generalization to Logical Reasoning}
\label{subsec:gdr}
In this section, we verify whether LLMs enhanced by \texttt{LLM-ERL} can be generalized to other tasks that need logical reasoning. We translate the symbolic representations of event relations into a form of deductive reasoning (i.e., containing facts, rules, and queries) to maintain consistency in task settings. The prompt example can be found in Appendix~\ref{app:prompt_deductive}.
\paragraph{Dataset Construction}
We conduct experiments on two datasets: ProofWriter~\citep{tafjord2020proofwriter} and FOLIO~\citep{han2022folio}. Details of the datasets can be found in Appendix~\ref{app:dataset}.
\begin{table}
    \renewcommand
    \arraystretch{1.0}
    \centering
    \small
    \setlength{\tabcolsep}{8pt}
        \begin{tabular}{ll|c|c}
        \toprule
        \multicolumn{2}{l|}{\bf{Model (\%)}}
         & {\textbf{ProofWriter}}  & {\textbf{FOLIO}}  \\ 
        \midrule
        \multirow{3}{*}{\bf{Vicuna}} &vanilla ICL & 37 / 38  & 40 / 43 \\
        & vanilla CoT &40 / 42  &  38 / 40\\
        & CoT w. logic. &42 / 44  &42 / 45\\
        \midrule
        \multirow{3}{*}{\bf{Llama2}} &vanilla ICL &29 / 33  &42 / 45\\
        & vanilla CoT &31 / 37 &44 / 46\\
        & CoT w. logic. &40 / 42 &46 / 48\\
        \bottomrule
        \end{tabular}
       \caption{ \label{tab:deductive} Vicuna and Llama2's performance on ProofWriter and FOLIO before and after fine-tuning on \texttt{LLM-ERL}~(split by ``/'').}
       \vspace{-4mm}
\end{table}
\paragraph{Results}
As shown in Table~\ref{tab:deductive}, we are surprised to find that models fine-tuned on \texttt{LLM-ERL} (e.g., Llama2-FT) can also bring performance improvement on other logical reasoning datasets, even though \texttt{LLM-ERL} focuses on event relation logic. This shows that the logical reasoning ability acquired by LLMs in the fine-tuning process can be generalized to other domains. We intend to explore this intriguing aspect in future work.
\begin{figure}
\centering  
\includegraphics[width=0.48\textwidth]{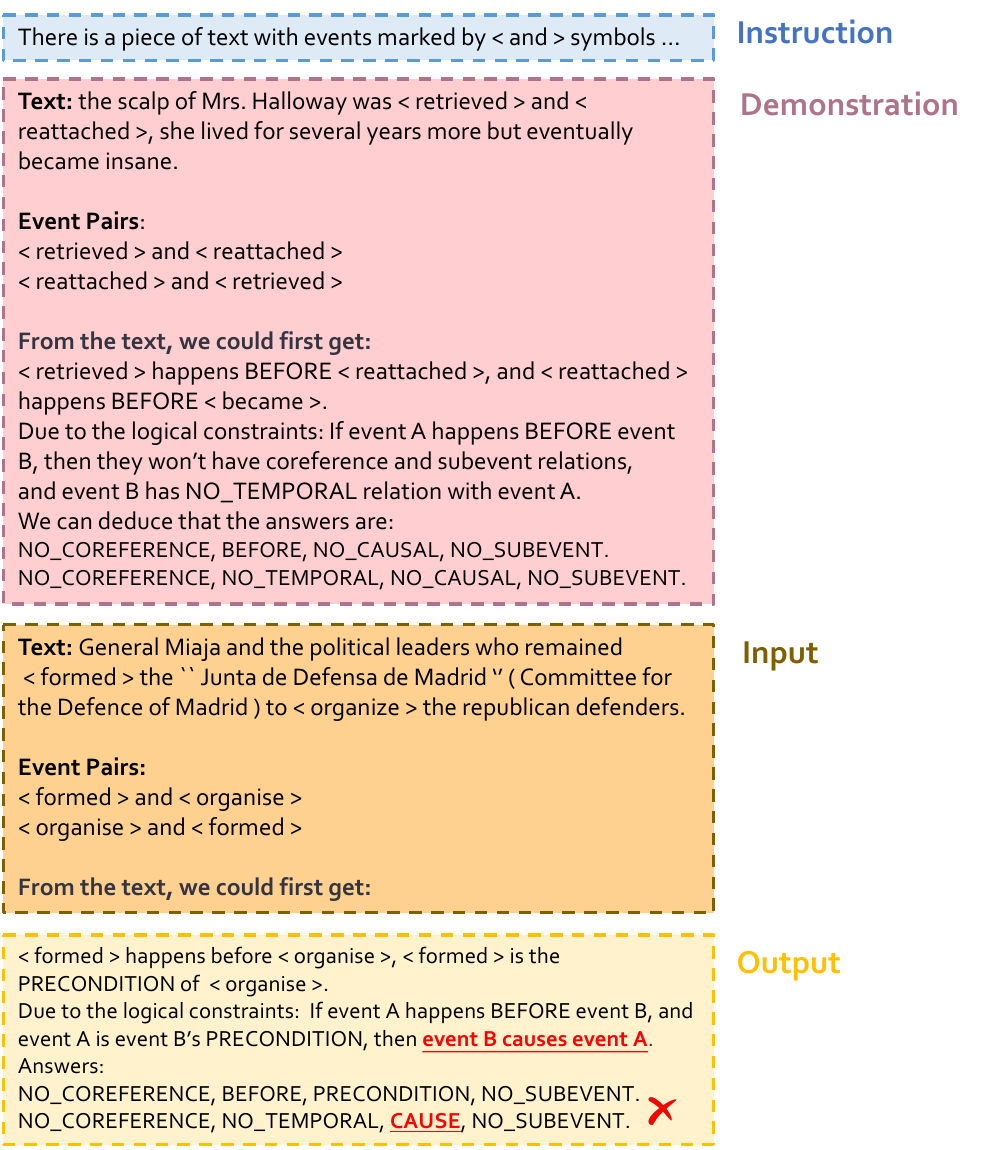}
\caption{A case study that ChatGPT generates inaccurate logical constraints.}
\label{fig:case}
\end{figure}

\begin{figure}
\centering 
    \includegraphics[width=0.48\textwidth]{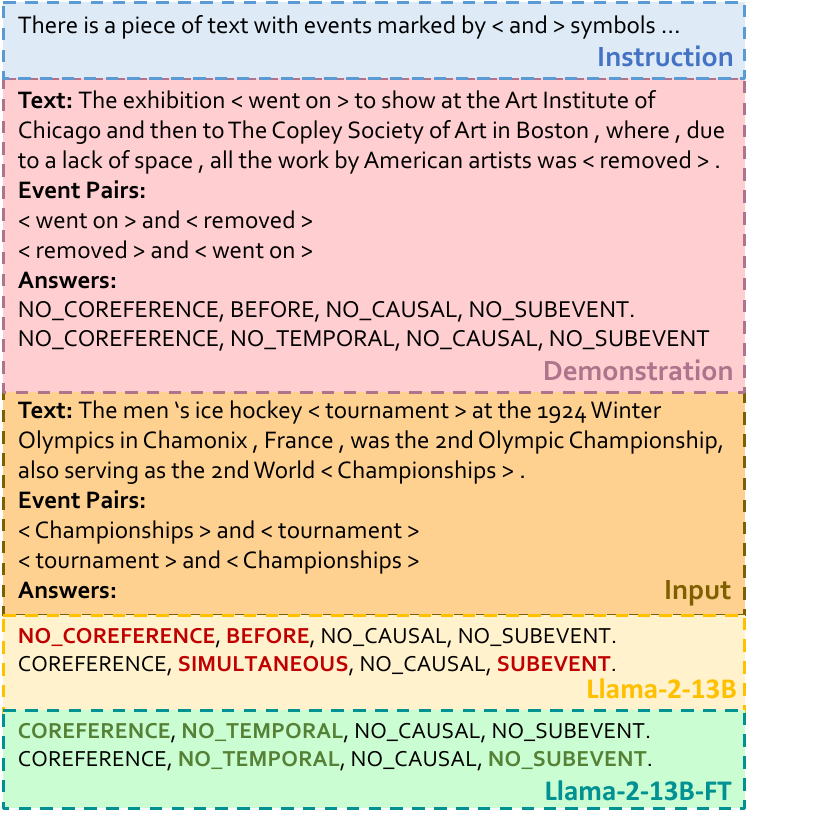}
  \caption{Case study on Llama-2-13B before and after fine-tuning (FT).}
  \label{fig:llama-case}
    \vspace{-4mm}
\end{figure}
\section{\textbf{Case Study}}

\subsection{Case Study on Self-generated Logical Constraints}
\label{app:case}
In the main context, we have found that directly
using CoT to infer logic does not help much for ERE tasks. One possible reason is that the inherent issues may lead to the failure of LLM in the precise rationale generation.
To further illustrate an intuitive impression, we conduct a case study on MAVEN-ERE and find that the logical constraints generated by LLMs themselves are often inaccurate in content.
As shown in Figure~\ref{fig:case}, ChatGPT could follow the logical constraint provided in the demonstration to a certain extent. However, it wrongly applies this to other relations --- knowing that event $A$ is event $B$'s precondition, it is wrong to think that event $B$ will cause event $A$. Actually, according to the logical constraints in Table~\ref{tab:lc_of_two}, the relations between ($B$, $A$) should be ``NO\_COREFERENCE, NO\_TEMPORAL, NO\_CAUSAL, NO\_SUBEVENT''.

\subsection{Case Study on Llama2 and Llama2-FT}
\label{app:llama-case}
In Figure~\ref{fig:llama-case}, We conduct a case study of Llama2-13B's answers to the same input before and after fine-tuning. From Figure~\ref{fig:llama-case} we can see that Llama2-FT could output the correct answers after fine-tuning on  \texttt{LLM-ERL},  which validates the effectiveness of our fine-tuning approach.

\begin{table*}[htbp!]
    \renewcommand
    \arraystretch{1.0}
    \centering
    \setlength{\tabcolsep}{7pt}
    \begin{tabular}{l|c}
    \toprule
         {\textbf{If Relation(A, B)}}  & \textbf{Prompt Text} \\ \midrule
         COREFERENCE & \makecell[c]{If event $A$ and event $B$ are COREFERENCE, \\ then they won’t have temporal, causal, and subevent relations,\\ and COREFERENCE relation is bidirectional.}  \\ \midrule
         NO\_TEMPORAL &\makecell[c]{If event $A$ and event $B$ do not have a temporal relation, \\then they won't have causal and subevent relations.} \\\midrule
         BEFORE &\makecell[c]{If event $A$ happens BEFORE event $B$,\\ then they won't have coreference and subevent relations,\\
         and event $B$ has NO\_TEMPORAL relation with event $A$.} \\\midrule
         OVERLAP &\makecell[c]{If event $A$ happens OVERLAP with event $B$, \\then they won't have coreference and subevent relations,\\
         and event $B$ has NO\_TEMPORAL relation with event $A$.} \\\midrule 
         CONTAINS &\makecell[c]{If event $A$’s time CONTAINS event $B$’s time, \\then they won't have coreference and causal relations,\\
         and event $B$ has NO\_TEMPORAL relation with event $A$.} \\\midrule
         SIMULTANEOUS & \makecell[c]{If event $A$ and event $B$ happen SIMULTANEOUSly, \\then they won’t have coreference, causal, and subevent relations,\\ and SIMULTANEOUS relation is bidirectional.} \\\midrule
         ENDS-ON &\makecell[c]{If event $A$ ENDS-ON event $B$, \\then they won’t have coreference, causal and subevent relations,\\
         and event $B$ has NO\_TEMPORAL relation with event $A$.} \\\midrule
         BEGINS-ON &\makecell[c]{If event $A$ BEGINS-ON event $B$, \\then they won’t have coreference, causal and subevent relations\\ and BEGINS-ON relation is bidirectional.} \\\midrule
         CAUSE &\makecell[c]{If event $A$ CAUSEs event $B$, \\then event $A$ happens BEFORE or OVERLAP event $B$, \\and they won't have coreference and subevent relations,\\
         and event $B$ has NO\_TEMPORAL relation with event $A$.} \\\midrule
         PRECONDITION &\makecell[c]{If event $A$ is event $B$’s PRECONDITION, \\then event $A$ happens BEFORE or OVERLAP event $B$,\\ and they won't have coreference and subevent relations,\\
         and event $B$ has NO\_TEMPORAL relation with event $A$.} \\\midrule
         SUBEVENT &\makecell[c]{If event $B$ is a SUBEVENT of event $A$, \\then they won’t have coreference and causal relations,\\ and event $A$’s time should CONTAINS event $B$’s time,\\
         and event $B$ has NO\_TEMPORAL relation with event $A$.} \\
    \bottomrule
    \end{tabular}
       \caption{\label{tab:lc_of_two_txt} Prompt text of relations between two events.}
\end{table*}
\begin{table*}[htbp!]
    \renewcommand
    \arraystretch{1.0}
    \centering
    \small
    \setlength{\tabcolsep}{5pt}
    \begin{tabular}{l|c}
    \toprule
         {\textbf{If Relation($A$, $B$) $\wedge$ Relation($B$, $C$)}}  & \textbf{Then Relation ($A$, $C$)} \\ \midrule
         COREFERENCE $\wedge$ COREFERENCE &COREFERENCE, $\neg$TEMPORAL, $\neg$CAUSAL,  $\neg$SUBEVENT   \\ 
         COREFERENCE $\wedge$ BEFORE &BEFORE, $\neg$COREFERENCE, $\neg$SUBEVENT \\
         COREFERENCE $\wedge$ OVERLAP &OVERLAP, $\neg$COREFERENCE, $\neg$SUBEVENT \\
         COREFERENCE $\wedge$ CONTAINS &CONTAINS, $\neg$COREFERENCE, $\neg$CAUSAL \\
         COREFERENCE $\wedge$ SIMULTANEOUS &SIMULTANEOUS, $\neg$COREFERENCE,  $\neg$CAUSAL,  $\neg$SUBEVENT \\
         COREFERENCE $\wedge$ ENDS-ON &ENDS-ON, $\neg$COREFERENCE, $\neg$CAUSAL, $\neg$SUBEVENT \\
         COREFERENCE $\wedge$ BEGINS-ON &BEGINS-ON, $\neg$COREFERENCE, $\neg$CAUSAL, $\neg$SUBEVENT \\
         COREFERENCE $\wedge$ CAUSE &CAUSE, $\neg$COREFERENCE, BEFORE $\vee$ OVERLAP,  $\neg$SUBEVENT \\
         COREFERENCE $\wedge$ PRECONDITION &PRECONDITION, $\neg$COREFERENCE, BEFORE $\vee$ OVERLAP, $\neg$SUBEVENT \\
         COREFERENCE $\wedge$ SUBEVENT &SUBEVENT, $\neg$COREFERENCE, CONTAINS $\neg$CAUSAL\\
         BEFORE $\wedge$ BEFORE &BEFORE, $\neg$COREFERENCE, $\neg$SUBEVENT \\
         BEFORE $\wedge$ OVERLAP &BEFORE, $\neg$COREFERENCE, $\neg$SUBEVENT \\
         BEFORE $\wedge$ CONTAINS &BEFORE, $\neg$COREFERENCE, $\neg$SUBEVENT \\
         BEFORE $\wedge$ SIMULTANEOUS &BEFORE, $\neg$COREFERENCE, $\neg$SUBEVENT \\
         BEFORE $\wedge$ ENDS-ON &BEFORE, $\neg$COREFERENCE, $\neg$SUBEVENT \\
         BEFORE $\wedge$ BEGINS-ON &BEFORE, $\neg$COREFERENCE, $\neg$SUBEVENT \\
         OVERLAP $\wedge$ BEFORE &BEFORE, $\neg$COREFERENCE, $\neg$SUBEVENT\\ 
         OVERLAP $\wedge$ SIMULTANEOUS &OVERLAP, $\neg$COREFERENCE, $\neg$SUBEVENT\\  
         CONTAINS $\wedge$ CONTAINS &CONTAINS, $\neg$COREFERENCE, $\neg$CAUSAL\\ 
         CONTAINS $\wedge$ SIMULTANEOUS &CONTAINS, $\neg$COREFERENCE, $\neg$CAUSAL\\
         SIMULTANEOUS $\wedge$ BEFORE &BEFORE, $\neg$COREFERENCE, $\neg$SUBEVENT \\
         SIMULTANEOUS $\wedge$ OVERLAP &OVERLAP, $\neg$COREFERENCE, $\neg$SUBEVENT \\
         SIMULTANEOUS $\wedge$ CONTAINS &CONTAINS, $\neg$COREFERENCE, $\neg$CAUSAL\\
         SIMULTANEOUS $\wedge$ SIMULTANEOUS &SIMULTANEOUS, $\neg$COREFERENCE,  $\neg$CAUSAL,  $\neg$SUBEVENT\\
         SIMULTANEOUS $\wedge$ ENDS-ON &ENDS-ON, $\neg$COREFERENCE, $\neg$SUBEVENT \\
         SIMULTANEOUS $\wedge$ BEGINS-ON &BEGINS-ON, $\neg$COREFERENCE, $\neg$SUBEVENT \\
         SIMULTANEOUS $\wedge$ COREFERENCE &SIMULTANEOUS, $\neg$COREFERENCE,  $\neg$CAUSAL, $\neg$SUBEVENT\\ 
         ENDS-ON $\wedge$ CONTAINS &BEFORE, $\neg$COREFERENCE, $\neg$SUBEVENT\\ 
         ENDS-ON $\wedge$ BEGINS-ON &ENDS-ON, $\neg$COREFERENCE, $\neg$CAUSAL, $\neg$SUBEVENT\\ 
         ENDS-ON $\wedge$ SIMULTANEOUS &ENDS-ON, $\neg$COREFERENCE, $\neg$CAUSAL, $\neg$SUBEVENT\\ 
         BEGINS-ON $\wedge$ SIMULTANEOUS &BEGINS-ON, $\neg$COREFERENCE, $\neg$CAUSAL, $\neg$SUBEVENT\\
         BEGINS-ON $\wedge$ BEGINS-ON &BEGINS-ON, $\neg$COREFERENCE, $\neg$CAUSAL, $\neg$SUBEVENT\\
         BEGINS-ON  $\wedge$ COREFERENCE &BEGINS-ON, $\neg$COREFERENCE, $\neg$CAUSAL, $\neg$SUBEVENT \\
         CAUSE $\wedge$ CAUSE &CAUSE, $\neg$COREFERENCE, BEFORE $\vee$ OVERLAP,  $\neg$SUBEVENT\\
         CAUSE $\wedge$ SUBEVENT &CAUSE, $\neg$COREFERENCE, BEFORE $\vee$ OVERLAP,  $\neg$SUBEVENT\\
         PRECONDITION  $\wedge$ CAUSE &CAUSE, $\neg$COREFERENCE, BEFORE $\vee$ OVERLAP,  $\neg$SUBEVENT\\
         PRECONDITION $\wedge$ PRECONDITION &PRECONDITION, $\neg$COREFERENCE, BEFORE $\vee$ OVERLAP,  $\neg$SUBEVENT\\
         PRECONDITION $\wedge$ SUBEVENT &PRECONDITION, $\neg$COREFERENCE, BEFORE $\vee$ OVERLAP,  $\neg$SUBEVENT\\
         SUBEVENT $\wedge$ SUBEVENT &SUBEVENT, $\neg$COREFERENCE, CONTAINS $\neg$CAUSAL\\
    \bottomrule
    \end{tabular}
       \caption{\label{tab:lc_of_three} Logical Constraints for the transitivity rules among three events, where $\wedge$ denotes "AND", $\neg$ denotes "NOT", $\vee$ denotes "OR".}
\end{table*}
\begin{table*}[htbp!]
    \renewcommand
    \arraystretch{1.0}
    \centering
    \small
    \setlength{\tabcolsep}{7pt}
    \begin{tabular}{l|c}
    \toprule
         {\textbf{If Relation($A$, $B$) $\wedge$ Relation($B$, $C$)}}  & \textbf{Prompt Text} \\ \midrule
         COREFERENCE $\wedge$ COREFERENCE & \multirow{10}{*}{\makecell[c]{If event $A$ and event $B$ are COREFERENCE, \\then the relations between event $B$ and event $C$ \\should be the same as that between event $A$ and event $C$.}}   \\ 
         COREFERENCE $\wedge$ BEFORE  \\
         COREFERENCE $\wedge$ OVERLAP  \\
         COREFERENCE $\wedge$ CONTAINS  \\
         COREFERENCE $\wedge$ SIMULTANEOUS \\
         COREFERENCE $\wedge$ ENDS-ON \\
         COREFERENCE $\wedge$ BEGINS-ON \\
         COREFERENCE $\wedge$ CAUSE  \\
         COREFERENCE $\wedge$ PRECONDITION \\
         COREFERENCE $\wedge$ SUBEVENT \\ \midrule
         BEFORE $\wedge$ BEFORE &\multirow{6}{*}{\makecell[c]{If event $A$ happens BEFORE event $B$, and Relation($B$, $C$), \\then event $A$ happens BEFORE event $C$.}} \\ 
         BEFORE $\wedge$ OVERLAP  \\
         BEFORE $\wedge$ CONTAINS  \\
         BEFORE $\wedge$ SIMULTANEOUS  \\
         BEFORE $\wedge$ ENDS-ON  \\
         BEFORE $\wedge$ BEGINS-ON  \\\midrule
         OVERLAP $\wedge$ BEFORE &\makecell[c]{If event $A$ happens OVERLAP with event $B$, \\and event $B$ happens BEFORE event $C$,\\ then event $A$ happens BEFORE event $C$.}\\ \midrule
         OVERLAP $\wedge$ SIMULTANEOUS &\makecell[c]{If event $A$ happens OVERLAP with event $B$, \\and event $B$ and event $C$ happen SIMULTANEOUSly,\\ then event $A$ happens BEFORE event $C$.}\\  \midrule
         CONTAINS $\wedge$ CONTAINS &\makecell[c]{If event $A$’s time CONTAINS event $B$’s time, \\and event $B$’s time CONTAINS event $C$’s time, \\then event $A$’s time CONTAINS event $C$’s time.}\\  \midrule
         CONTAINS $\wedge$ SIMULTANEOUS &\makecell[c]{If event $A$’s time CONTAINS event $B$’s time, \\and event $B$ and event $C$ happen SIMULTANEOUSly,\\ then event $A$’s time CONTAINS event $C$’s time.}\\  \midrule
         SIMULTANEOUS $\wedge$ BEFORE  &\multirow{6}{*}{\makecell[c]{If events A and B happen SIMULTANEOUSly, and Relation($B$, $C$), \\ then event $A$'s time CONTAINS event $C$'s time.}} \\
         SIMULTANEOUS $\wedge$ OVERLAP  \\
         SIMULTANEOUS $\wedge$ CONTAINS \\
         SIMULTANEOUS $\wedge$ SIMULTANEOUS \\
         SIMULTANEOUS $\wedge$ ENDS-ON  \\
         SIMULTANEOUS $\wedge$ BEGINS-ON \\\midrule
         ENDS-ON $\wedge$ CONTAINS &\makecell[c]{If event $A$ ENDS-ON event $B$, \\and event $B$’s time CONTAINS event $C$’s time, \\then event $A$ happens BEFORE event $C$.}\\  \midrule
         ENDS-ON $\wedge$ BEGINS-ON &\multirow{2}{*}{\makecell[c]{If event $A$ ENDS-ON event $B$, and Relation($B$, $C$), \\ then event $A$ ENDS-ON event $C$.}} \\
         ENDS-ON $\wedge$ SIMULTANEOUS \\ \midrule
         BEGINS-ON $\wedge$ SIMULTANEOUS &\multirow{2}{*}{\makecell[c]{If event $A$ BEGINS-ON event $B$, and Relation($B$, $C$),\\ then event $A$ BEGINS-ON event $C$.}} \\
         BEGINS-ON $\wedge$ BEGINS-ON \\\midrule
         CAUSE $\wedge$ CAUSE &\makecell[c]{If event $A$ CAUSEs event $B$,\\ and event $B$ CAUSEs event $C$, \\then event $A$ CAUSEs event $C$.}\\  \midrule
         CAUSE $\wedge$ PRECONDITION &\makecell[c]{If event $A$ CAUSEs event $B$,\\ and event $B$ is event $C$’s PRECONDITION, \\then event $A$ is event $C$’s PRECONDITION.}\\  \midrule
         CAUSE $\wedge$ SUBEVENT&\makecell[c]{If event $A$ CAUSEs event $B$, \\and event $C$ is a SUBEVENT of event $B$, \\then event $A$ CAUSEs event $C$.}\\  \midrule
         PRECONDITION $\wedge$ PRECONDITION &\makecell[c]{If event $A$ is event $B$’s PRECONDITION, \\and event $B$ is event $C$’s PRECONDITION, \\then event $A$ is event $C$’s PRECONDITION.}\\  \midrule
         PRECONDITION $\wedge$ SUBEVENT &\makecell[c]{If event $A$ is event $B$’s PRECONDITION, \\and event $C$ is a SUBEVENT of event $B$, \\then event $A$ is event $C$’s PRECONDITION.}\\  \midrule
         SUBEVENT $\wedge$ SUBEVENT &\makecell[c]{If event $B$ is a SUBEVENT of event $A$, \\and event $C$ is a SUBEVENT of event $B$, \\then event $C$ is a SUBEVENT of event $A$.}\\ 
    \bottomrule
    \end{tabular}
       \caption{\label{tab:lc_of_three_txt} Prompt text of relations among three events.}
\end{table*}
\clearpage

\section{Prompt Examples}

In this section, we provide examples of prompts used for each task and approach. 

\subsection{Pilot Case Study}
\label{app:prompt_pilot}
In the context of our paper, ``relevant logical constraints'' refer to the necessary knowledge or requirements for processing the current sample. They are accurately defined and closely related to the case in question. On the other hand, ``irrelevant logical constraints'' denote logic that, while possibly correct in content, does not directly pertain to the specific sample at hand. This distinction is crucial to maintain the focus and relevance of our analysis. 

\begin{figure*}[htbp!]
\centering  
\includegraphics[width=0.9\textwidth]{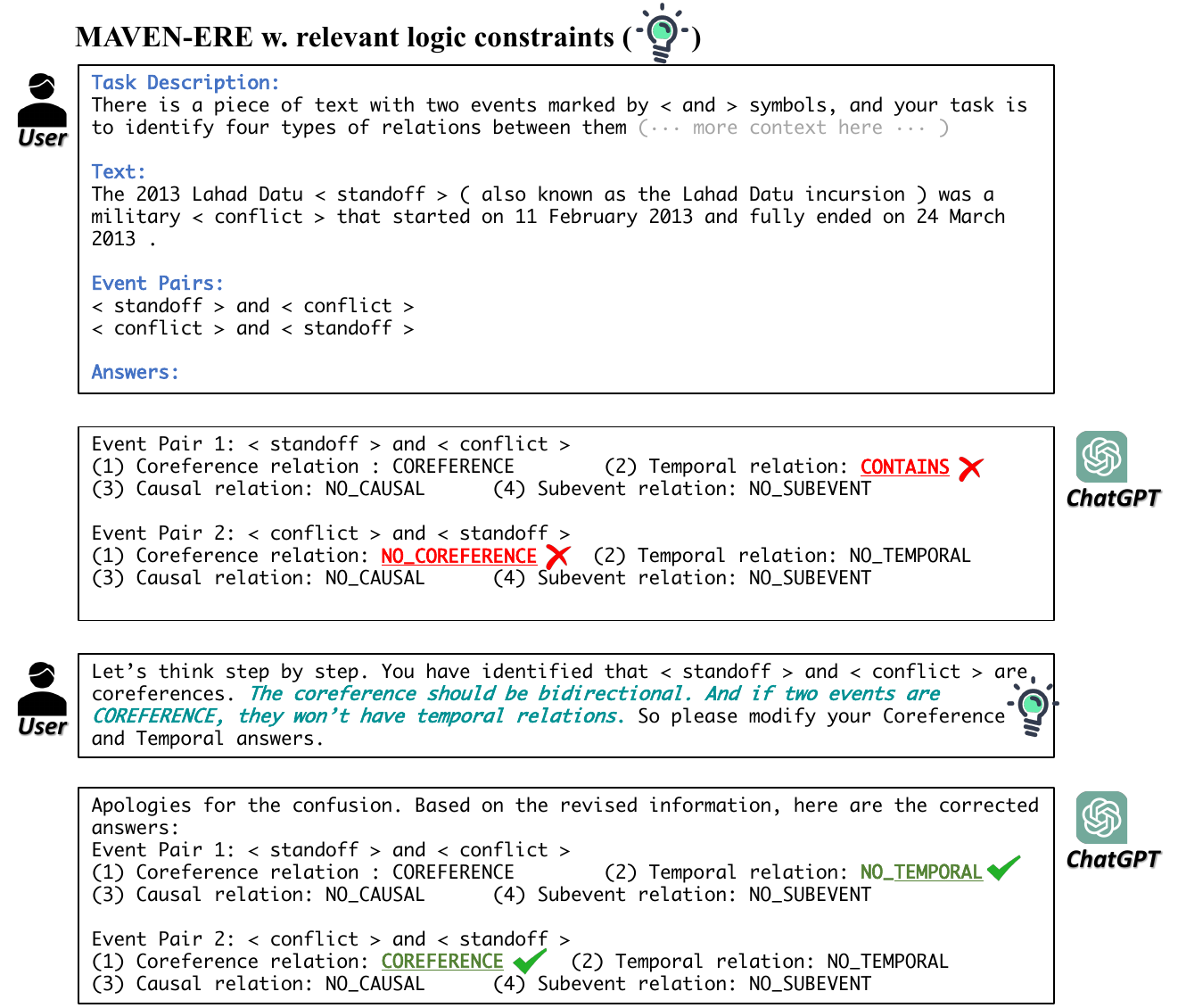}
\caption{MAVEN-ERE w. relevant logic constraints}
\label{fig:prompt_pilot1}
\end{figure*}

\begin{figure*}[htbp!]
\centering  
\includegraphics[width=1.0\textwidth]{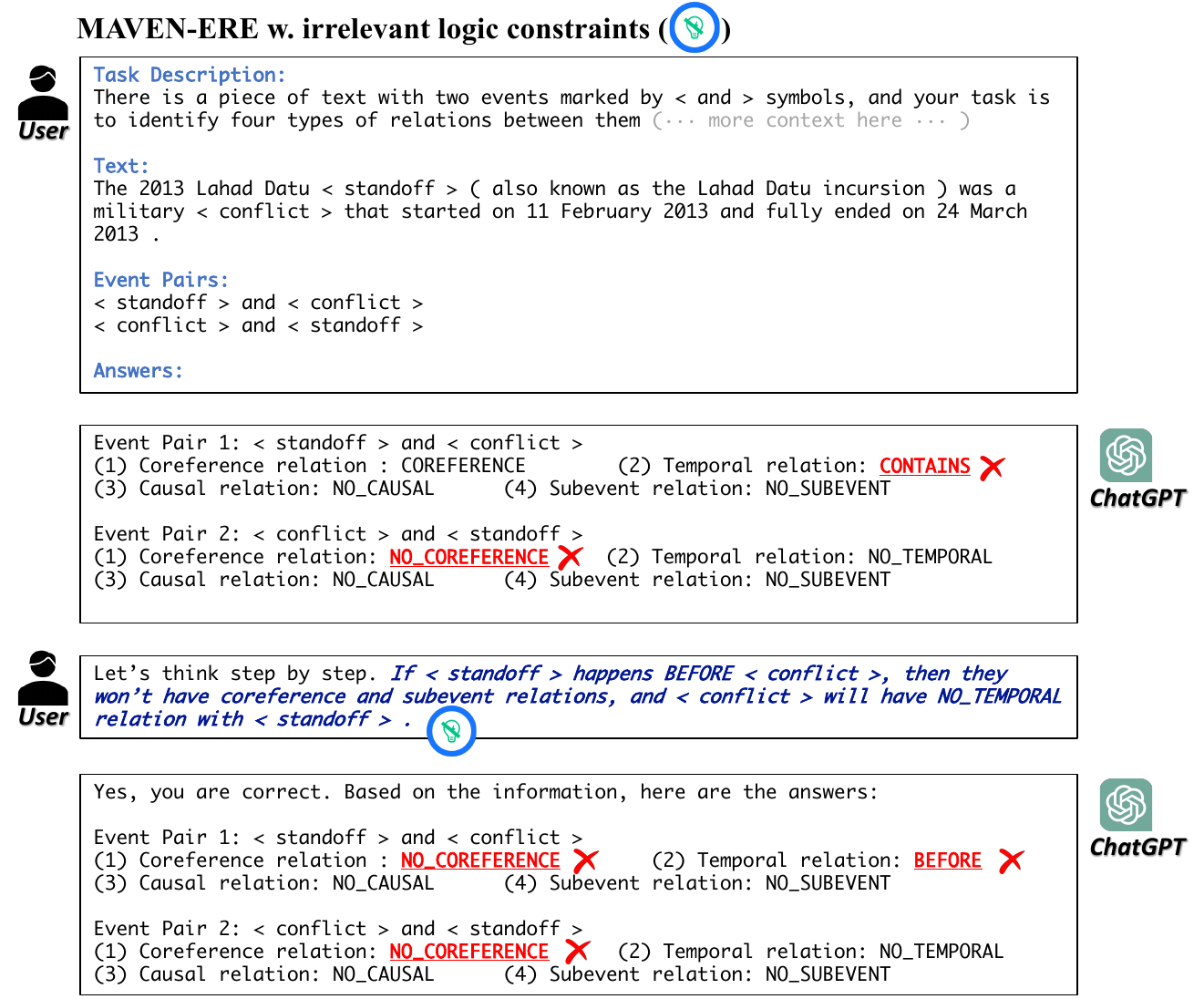}
\caption{MAVEN-ERE w. irrelevant logic constraints}\label{fig:prompt_pilot2}
\end{figure*}

\begin{figure*}[htbp!]
\centering  
\includegraphics[width=1.0\textwidth]{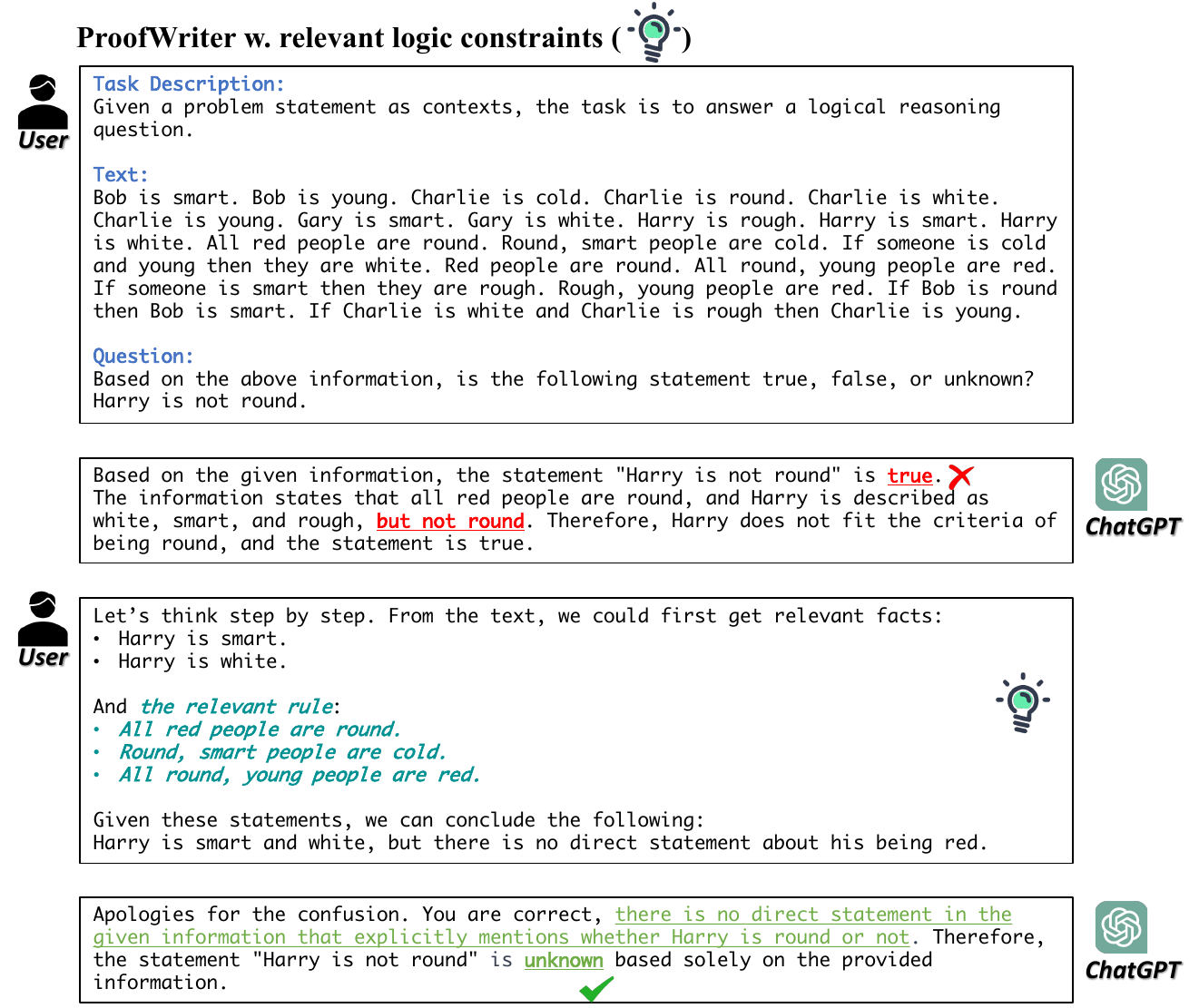}
\caption{ProofWriter w. relevant logic constraints}\label{fig:prompt_pilot3}
\end{figure*}

\begin{figure*}[htbp!]
\centering  
\includegraphics[width=1.0\textwidth]{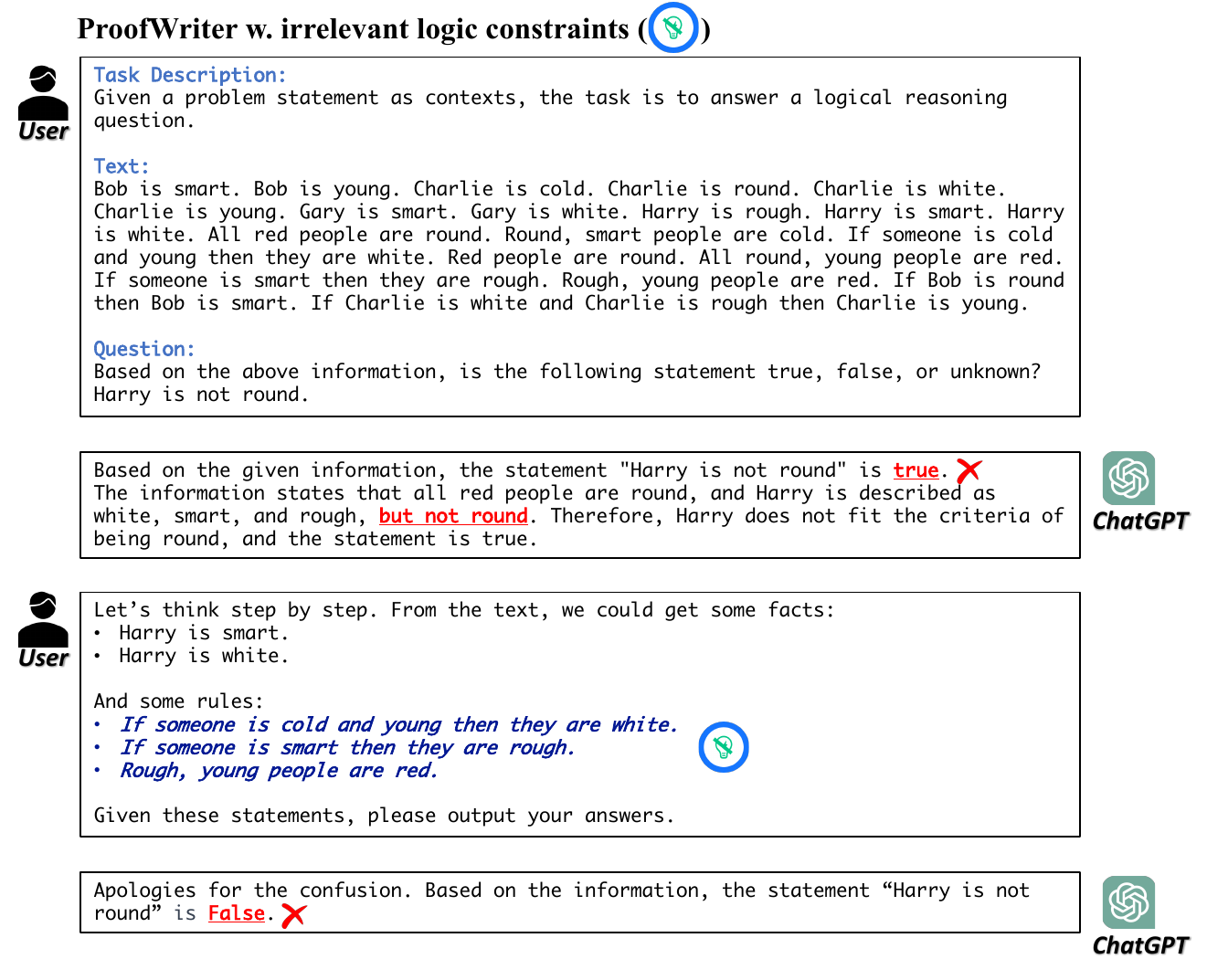}
\caption{ProofWriter w. irrelevant logic constraints}\label{fig:prompt_pilot4}
\end{figure*}

\paragraph{Process of Determining Relevant Logic}
\begin{itemize}
    \item
For MAVEN-ERE: we have presented the critical importance of ensuring the logical consistency of answers generated by LLMs. Therefore, we implement a rigorous manual check of the LLM outputs. During this process, we specifically identify and rectify any logical inconsistencies. 
We guide LLM by incorporating the most relevant logical constraints from Table~\ref{tab:lc_of_two_txt} into the LLM's instruction, thereby facilitating the refinement and accuracy of its responses.
\item
For ProofWriter: we have observed that the context often contains some facts and rules that are not directly pertinent to the current question. Therefore, we start by analyzing the question at hand and the initial answers provided by the LLM. Based on this, we selectively introduce rules and facts that are specifically relevant to the current scenario. This method allows us to provide the LLM with focused guidance, enabling it to refine its answers more effectively and accurately.
\end{itemize}
\paragraph{Process of Determining Irrelevant Logic}
\begin{itemize}
    \item
For MAVEN-ERE: We randomly sample 1-2 constraints from the entire set removing those relevant logical constraints and construct the prompts based on each sample.
\item
For ProofWriter: We artificially select irrelevant logical constraints from each sample’s content, thereby introducing a form of ``noise'' or ``distraction'' to the LLM's judgment process.
\end{itemize}
\paragraph{Prompt Examples}
\begin{itemize}
    \item MAVEN-ERE w. relevant logic constraints (Figure~\ref{fig:prompt_pilot1});
    \item MAVEN-ERE w. irrelevant logic constraints (Figure~\ref{fig:prompt_pilot2});
    \item ProofWriter w. relevant logic constraints (Figure~\ref{fig:prompt_pilot3});
    \item ProofWriter w. irrelevant logic constraints (Figure~\ref{fig:prompt_pilot4}).

\end{itemize}

\clearpage
\clearpage
\clearpage

\begin{figure*}
\begin{minipage}{\linewidth}
  \subsection{Incoporating Logical Constraints}
\label{app:prompt_lc}
The \sethlcolor{lightgreen}\hl{highlighted parts} represent the content generated by LLMs. We omit the demonstration here for clarity.
\\
\\
  \includegraphics[width=0.95\textwidth]{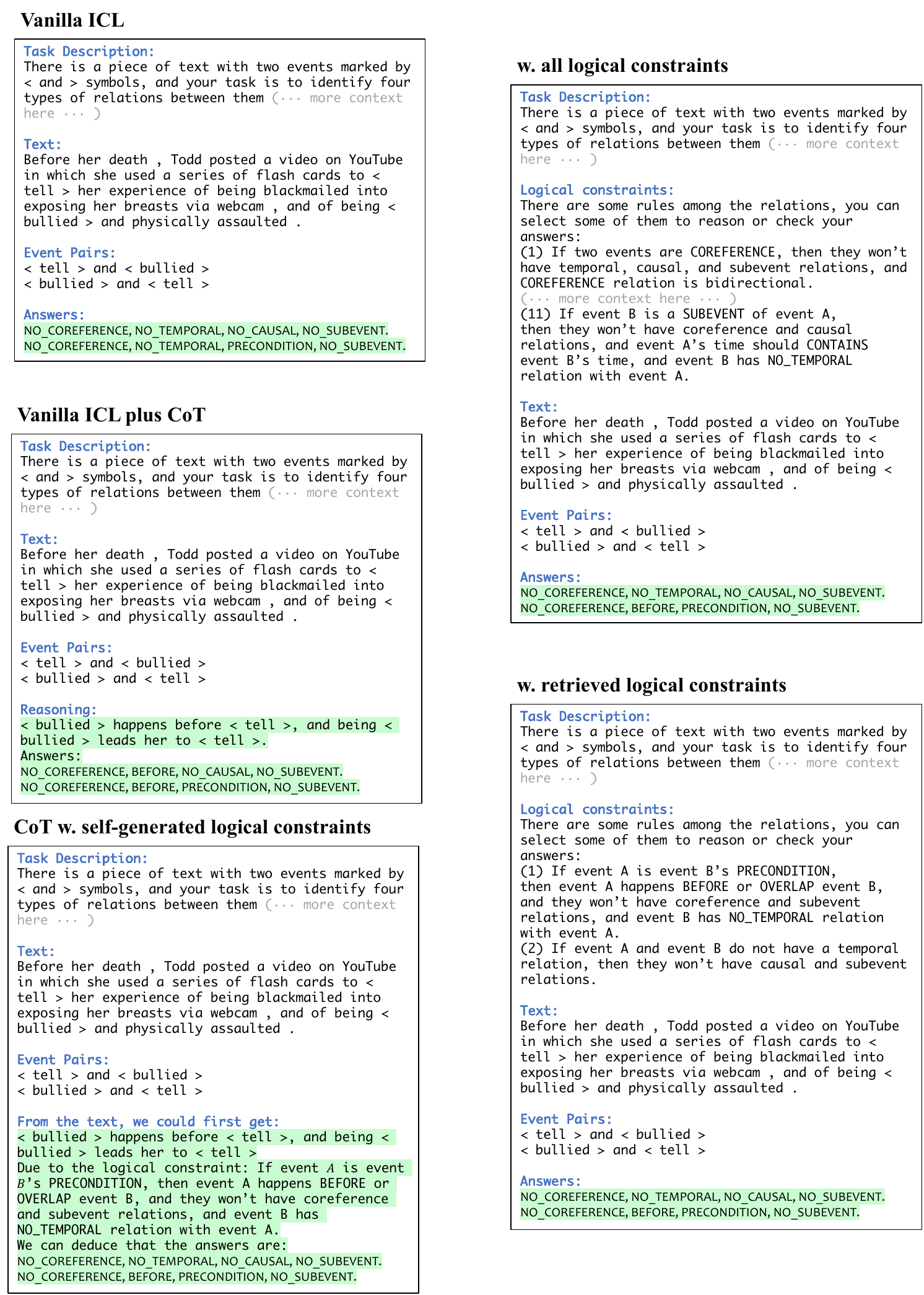}
  \label{fig:prompt_lc}
\end{minipage}
\end{figure*}

\clearpage

\begin{figure*}
\begin{minipage}{\linewidth}

\subsection{Iterative Retrievals}
\label{app:prompt_iterative}
In this section, we present a prompt example used in Section~\ref{subsec:ablation}. As shown in Figure~\ref{fig:prompt_iteration}, with iterative prompting, ChatGPT finally outputs the correct answers.
\\
\\
\includegraphics[width=0.8\textwidth]{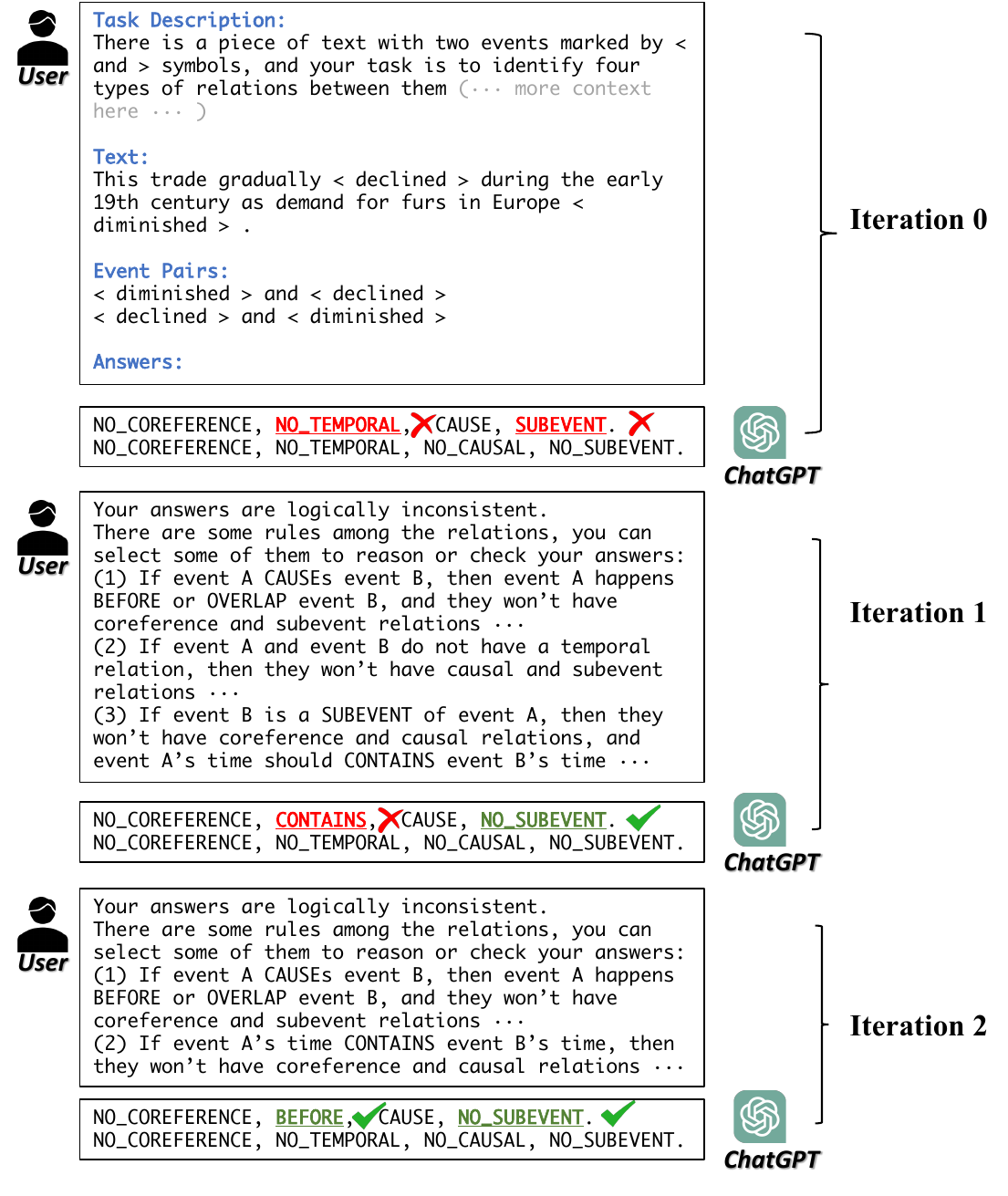}
\caption{Multi-turn conversation with ChatGPT. We retrieve relevant logical constraints and provide them to ChatGPT.}
\label{fig:prompt_iteration}
\end{minipage}
\end{figure*}
\clearpage

\begin{figure*}
\begin{minipage}{\linewidth}
  \subsection{Deductive Reasoning}
   \label{app:prompt_deductive}
The \sethlcolor{lightgreen}\hl{highlighted parts} represent the content generated by LLMs. We omit the demonstration here for clarity.

  \includegraphics[width=1.0\textwidth]{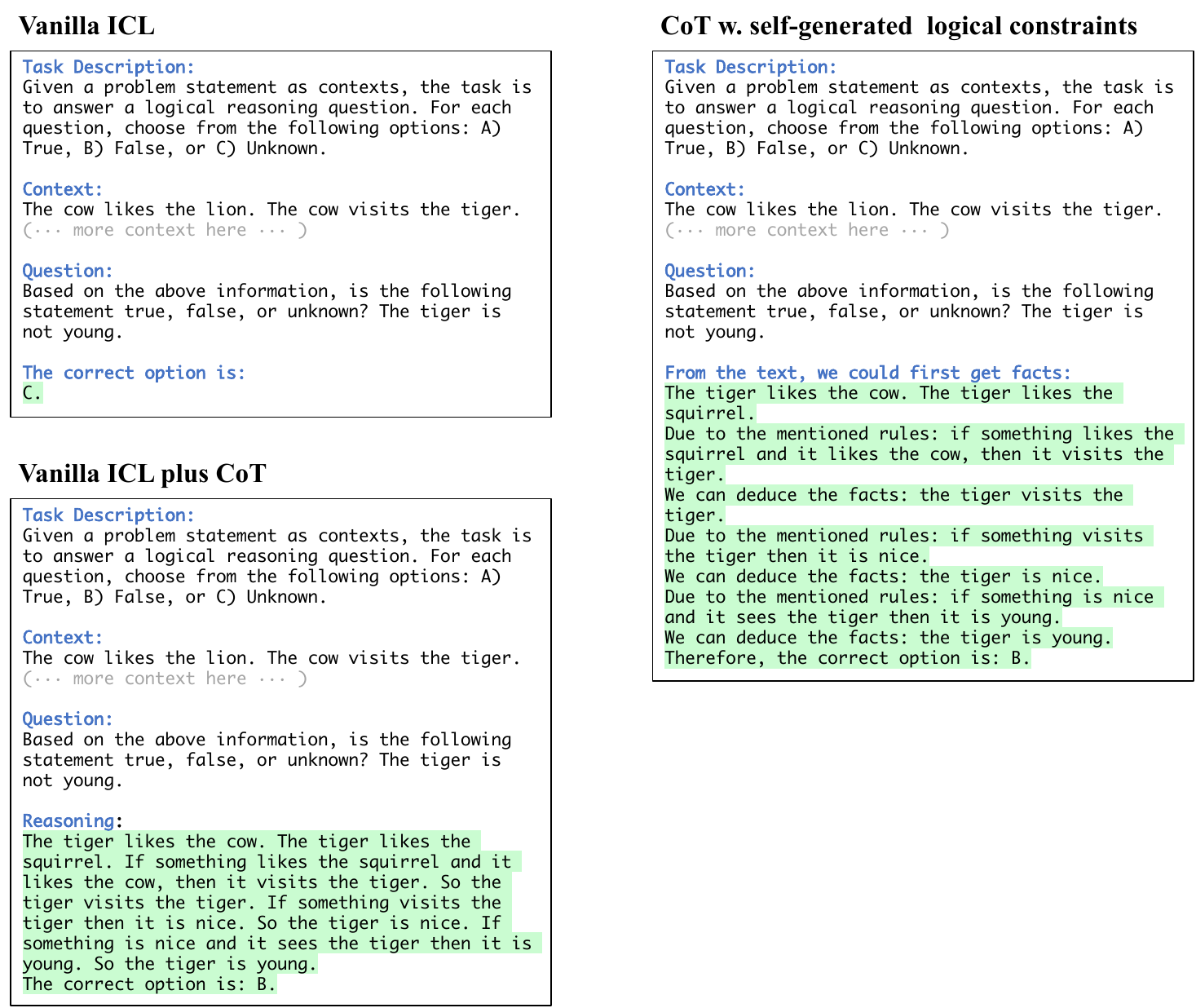}
\label{fig:prompt_deductive}
\end{minipage}
\end{figure*}

\end{document}